\PassOptionsToPackage{dvipsnames,svgnames,table}{xcolor}
\documentclass[10pt,journal,compsoc]{IEEEtran}
\usepackage[numbers,square,comma]{natbib}

\usepackage[utf8]{inputenc} 
\usepackage[T1]{fontenc}    
\usepackage{hyperref}       
\usepackage{url}            
\usepackage{booktabs}       
\usepackage{stfloats}
\usepackage{amsfonts}       
\usepackage{nicefrac}       
\usepackage{microtype}      
\usepackage{xcolor}         
\usepackage{graphicx}
\usepackage{subfigure}
\usepackage{amsmath,amssymb} 
\usepackage{caption} 

\usepackage{wrapfig}
\usepackage[american]{babel}

\usepackage{times}
\usepackage{forest}
\usepackage{array,tabularx}
\newcolumntype{M}[1]{>{\raggedright\arraybackslash}m{#1}}
\newcommand{\comparisonpanel}[1]{%
  \par\addvspace{6pt}{\raggedright\bfseries #1\par}\nobreak\vspace{2pt}}
\hypersetup{pdftitle={Understanding Diffusion Acceleration: Methods, Trade-Offs, and Open Questions}}
\newcolumntype{C}{c}
\usepackage{amsthm}
\usepackage{multirow}
\usepackage{bm}
\usepackage{overpic}
\usepackage[switch]{lineno}  %

\usepackage{threeparttable}
\usepackage{tikz}
\usepackage{pifont}

\definecolor{DarkGreen}{RGB}{1,50,32}
\definecolor{dr}{RGB}{47,85,151}
\definecolor{dreg}{RGB}{112,48,160}
\definecolor{dn}{RGB}{255, 117, 143}
\definecolor{pr}{RGB}{228,197,111}
\definecolor{kd}{RGB}{251,133,0}
\definecolor{mr}{RGB}{84,130,53}
\definecolor{tbs}{RGB}{2,48,71}

\usepackage{placeins}
\begin{document}
\title{Understanding Diffusion Acceleration:\\\smash[b]{Methods, Trade-Offs, and Open Questions}}
\IEEEtitleabstractindextext{%
\begin{abstract}
Diffusion and flow-based generative models support high-quality image, audio, and video synthesis. Sampling can require dozens to hundreds of denoiser evaluations, with each pass becoming more expensive as resolution, video length, or model size increases. This cost affects latency and energy use in interactive and edge applications. We examine four approaches to accelerating inference: distillation, consistency training, sampling methods, and denoiser design. We express training objectives, transition maps, and numerical updates in a common notation to compare changes to the learning target, the sampling trajectory, and the work performed within each denoiser pass. Published image and video benchmarks illustrate the quality--cost trade-offs of one- and few-step generation, the interaction between solvers and timestep schedules, the effect of feature caching, and the difference between first-frame latency and sustained streaming throughput. We close with open problems in visual-native architectures and post-compression distillation, solver-aware few-step training, and semantic trajectory alignment with factorized distribution matching. Github: \textcolor{blue}{https://github.com/LynnReal-AI/awesome-diffusion-acceleration}.
\end{abstract}

\begin{IEEEkeywords}
Diffusion Models, Stochastic Differential Equation, Generative Adversarial Networks, Flow Model
\end{IEEEkeywords}}

\author{
Xiaofeng Mao$^{1,2\ast}$~~~Fu-yun Wang$^{3\ast}$~~~Le Zhuo$^{4\ast}$~~~Zhengkai Jiang$^{5\ast}$\\~~~Shaohao Rui$^{1,2\ast}$~~~Weijie Ma$^{1,2\ddagger}$\\
$^1$\normalfont{LynnReal AI}~~~$^2$\normalfont{Shanghai Innovation Institute}~~~\\
$^3$\normalfont{The Chinese University of Hong Kong}~~~$^4$\normalfont{Shanghai AI Laboratory} \\
$^5$\normalfont{Hong Kong University of Science and Technology}\\
}
\maketitle

\if TT\insert\footins{\noindent\footnotesize\\
    $^{\ast}$~Equal contribution. \quad $^{\ddagger}$~Corresponding author. \\
}\fi
\IEEEdisplaynontitleabstractindextext
\IEEEpeerreviewmaketitle
\enlargethispage{-2pt}

\definecolor{hidden-draw}{RGB}{0,0,0}    
\definecolor{coral}{RGB}{255, 127, 80}   
\definecolor{sage}{RGB}{128, 177, 133}   
\definecolor{azure}{RGB}{91, 155, 213}   
\definecolor{lilac}{RGB}{180, 160, 207}  
\tikzstyle{leaf}=[draw=hidden-draw,
    rounded corners,minimum height=1.2em,
    fill=hidden-orange!40,text opacity=1,    align=center,
    fill opacity=.5,  text=black,align=left,font=\scriptsize,
inner xsep=3pt,
inner ysep=1pt,
]

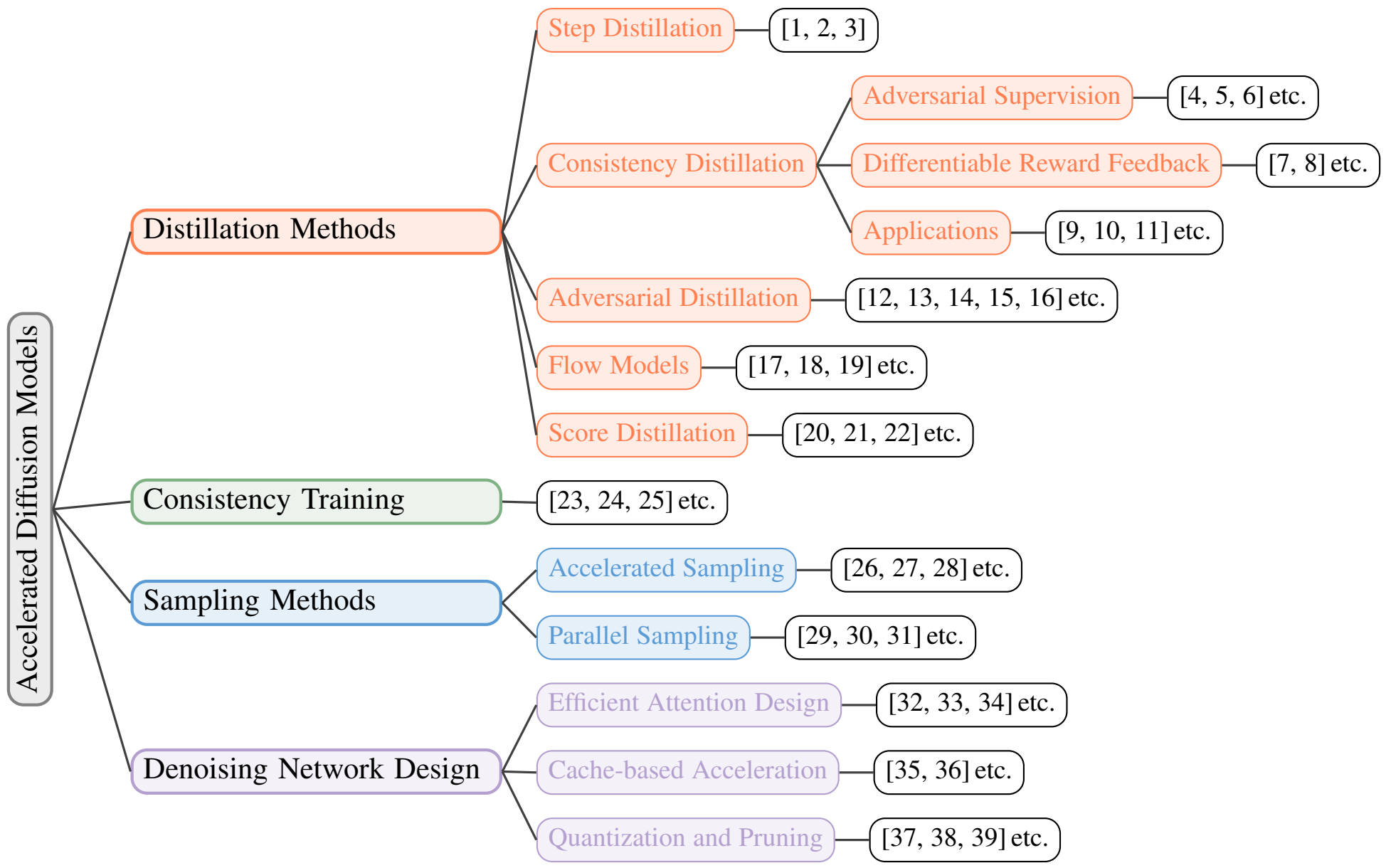
\begin{figure*}[thp]
\centering
\begin{forest}
    for tree={
        grow=east,
        reversed=true,
        anchor=base west,
        parent anchor=east,
        child anchor=west,
        base=left,
        font=\footnotesize,
        rectangle,
        draw,
        rounded corners,align=left,
        minimum width=2.5em,
        minimum height=1.2em,
        edge+={darkgray, line width=0.6pt},
        s sep=6pt,
        l sep=9pt,
        inner xsep=3pt,
        inner ysep=1pt,
        ver/.style={rotate=90, child anchor=north, parent anchor=south, anchor=center},
    },
    where level=1{text width=9.6em}{},
    where level=2{font=\scriptsize}{},
    where level=3{font=\scriptsize}{},
    where level=4{font=\scriptsize}{},
    where level=5{font=\scriptsize}{},
    [Accelerated Diffusion Models, draw=gray, color=gray!100, fill=gray!15, thick, text=black, ver
        [Distillation Methods, color=coral, fill=coral!15, thick, text=black
            [Step Distillation, color=coral, fill=coral!15
                    [
                        \cite{salimans2022progressive,meng2023distillation,berthelot2023tract}
                    ]
            ]
            [Consistency Distillation, color=coral, fill=coral!15
                [Adversarial Supervision,color=coral, fill=coral!15
                    [
                        \cite{liu2024scott,kim2023consistency,wang2024phased}\,etc.
                    ]
                ]
                [Differentiable Reward Feedback, color=coral, fill=coral!15
                    [
                        \cite{li2024t2v,li2024t2v2}\,etc.
                    ]
                ]
                [Applications, color=coral, fill=coral!15
                    [
                        \cite{wang2023videolcm,wang2024animatelcm,xiao2024ccm}\,etc.
                    ]
                ]
            ]
            [Adversarial Distillation, color=coral, fill=coral!15
                [
                    \cite{lin2024sdxl,sauer2025adversarial,sauer2024fast,mao2024osv,xie2024addsr}\,etc.
                ]
            ]
            [Flow Models, color=coral, fill=coral!15
                [
                    \cite{liu2022flow,liu2023instaflow,yang2024consistency}\,etc.
                ]
            ]
            [Score Distillation, color=coral, fill=coral!15
                [
                    \cite{nguyen2024swiftbrush,zhou2024score,zhou2024adversarial}\,etc.
                ]
            ]
        ]
        [Consistency Training, color=sage, fill=sage!15, thick, text=black
            [
                \cite{song2023consistency,lee2024truncated,geng2024consistency}\,etc.
            ]
        ]
        [Sampling Methods, color=azure, fill=azure!15, thick, text=black
            [Accelerated Sampling, color=azure, fill=azure!15
                [
                    \cite{lu2022dpm1,zhao2024unipc,zhou2024fast}\,etc.
                ]
            ]
            [Parallel Sampling, color=azure, fill=azure!15
                [
                    \cite{zheng2023fast,shih2024parallel,selvam2024self}\,etc.
                ]
            ]
        ]
        [Denoising Network Design, color=lilac, fill=lilac!15, thick, text=black
            [Efficient Attention Design, color=lilac, fill=lilac!15
                [
                        \cite{fei2024diffusion,tian2024u,teng2024dim}\,etc.
                ]
            ]
            [Cache-based Acceleration, color=lilac, fill=lilac!15
                [
                    \cite{ma2024deepcache,li2023faster}\,etc.
                ]
            ]
            [Quantization and Pruning, color=lilac, fill=lilac!15
                [
                    \cite{yang20241,castells2024ld,li2023qdiffusion}\,etc.
                ]
            ]
        ]
    ]
\end{forest}
\caption{Taxonomy of diffusion acceleration: Distillation, Consistency Training, Sampling Methods, and Denoising Network Design.}
\label{fig:acceleration-taxonomy}
\end{figure*}

\section{Introduction} \label{sec:intro}
\IEEEPARstart{D}{iffusion} Models (DMs)~\cite{croitoru2023diffusion,yang2023diffusion,ho2020denoising,song2020denoising,karras2022elucidating,dhariwal2021diffusion,nichol2021improved,peebles2023scalable,mao2026packforcing,mao2025yume,mao2026yume1,mao2026lynnreal} generate data by iteratively reversing a gradual corruption process, replacing Generative Adversarial Networks (GANs)~\cite{goodfellow2020generative,creswell2018generative,mao2017least} as the dominant paradigm for high-fidelity synthesis. Compared with adversarial training, DMs provide three practical advantages: (1) stable regression-based optimization, (2) broad mode coverage that avoids severe mode collapse, and (3) flexible conditioning for controllable generation. These properties have enabled scaling to billion-parameter backbones such as the 12-billion-parameter FLUX.1~\cite{flux2024}, driving progress across image synthesis~\cite{esser2024scaling,peebles2023scalable,chen2024pixart,gao2024lumina}, image super-resolution~\cite{xie2024addsr,yue2024resshift,yu2024scaling}, video synthesis~\cite{blattmann2023align,zhang2023i2vgen,yang2024cogvideox,rui2026biwm}, text generation~\cite{li2022diffusion,li2023diffusion}, speech synthesis~\cite{kong2020diffwave,liu2023audioldm}, anomaly detection~\cite{wolleb2022diffusion}, and AI4Science~\cite{guo2024diffusion,abramson2024accurate}.

Despite this generative quality, iterative sampling imposes a steep inference cost. Whereas GANs synthesize an output in a single forward pass, the original DDPM discretizes the reverse chain into 1,000 denoising steps~\cite{ho2020denoising}. Deterministic ODE samplers such as DDIM~\cite{song2020denoising} and higher-order solvers~\cite{lu2022dpm} reduce this budget to tens of evaluations, yet every retained step still runs a full pass through a large U-Net or Transformer denoiser. Pushing standard samplers into the single-digit step regime introduces pronounced discretization error and visual degradation, leaving inference speed far behind single-step generators and motivating dedicated acceleration techniques~\cite{salimans2022progressive,song2023consistency,lu2022dpm}.

Beyond raw latency, repeated inference dominates the operational energy footprint of deployed generative models. Earlier efficiency analyses centered largely on one-time training costs, whereas production services execute millions of sampling trajectories over a model's lifetime. Luccioni et al.~\cite{luccioni2024power} measure deployment energy across machine-learning workloads and identify image generation as one of the most power-intensive inference tasks. As modern text-to-image and text-to-video pipelines adopt large Diffusion Transformer (DiT) backbones~\cite{esser2024scaling,chen2024pixart,gao2024lumina}, end-to-end energy and compute compound across prompt encoding, iterative denoiser calls, and high-resolution latent decoding.

This computational bottleneck directly limits real-world deployment at both ends of the hardware spectrum. In video synthesis, 3D spatiotemporal attention scales quadratically with the joint number of spatial patches and frames, causing large video diffusion transformers~\cite{yang2024cogvideox} to take minutes per clip on datacenter GPUs and precluding interactive editing or live streaming. Conversely, deploying image generators on mobile and edge devices imposes strict memory and latency ceilings that require compact architectures and aggressive step reduction, as targeted by systems such as SnapFusion~\cite{li2024snapfusion}.

Prior surveys review general diffusion theory and applications~\cite{croitoru2023diffusion,yang2023diffusion}, focus on early knowledge distillation~\cite{luo2023comprehensive}, or address broader model compression and video-specific efficiency~\cite{ma2025efficientdiffusion,shao2026efficientvideo}. However, rapid advances across consistency distillation, flow-map training, distribution matching, learned solvers, and runtime caching~\cite{song2023consistency,song2023improved,wang2024phased,yin2024improved,sauer2025adversarial,geng2024consistency} have introduced diverse parameterizations and time conventions that make direct comparison difficult. We express training objectives and update rules in a shared notation (Table~\ref{symbols}) and pair this mathematical synthesis with published quantitative comparisons across image and streaming video benchmarks. The \href{https://github.com/LynnReal-AI/awesome-diffusion-acceleration}{companion repository} indexes the literature. We organize diffusion acceleration into four categories (Figure~\ref{fig:acceleration-taxonomy}):

\begin{table*}[!b]
\caption{Notation used in the main text and method tables.}
\centering
\small
\setlength{\tabcolsep}{5pt}
\renewcommand{\arraystretch}{1.04}
\begin{tabularx}{\linewidth}{@{}l>{\raggedright\arraybackslash}X@{}}
\toprule
\textbf{Main symbols} & \textbf{Symbol description} \\
\midrule
$\alpha_t,\sigma_t$ & Signal and noise scales: $x_t=\alpha_t x_0+\sigma_t\epsilon$, $q(x_t\mid x_0)=\mathcal N(\alpha_t x_0,\sigma_t^2I)$.\\
$t_n,N$ & Diffusion grid $t_0<t_1<\cdots<t_N$ with $N$ sampling intervals; generation runs from $t_N$ to $t_0$.\\
$x_t,\widehat{x}_s^{\phi}$ & State at time $t$ and a teacher-solver estimate at the earlier diffusion time $s$.\\
$\theta,\phi,\psi$ & Trainable student/model parameters, frozen reference parameters, and auxiliary score/velocity parameters.\\
$\theta^-,\bar\theta,\operatorname{sg}$ & Target weights (EMA or copied) and stop-gradient; iCT uses current weights in its stopped target.\\
$f_\theta$ & Consistency map; phase-conditioned maps also specify their endpoint $b$.\\
\addlinespace[2pt]
$\epsilon,\epsilon_\theta$ & Standard Gaussian perturbation and predicted noise.\\
$s_\theta,D_\theta$ & Score $s_\theta=-\epsilon_\theta/\sigma_t$ and clean-data predictor $D_\theta=(x_t-\sigma_t\epsilon_\theta)/\alpha_t$.\\
$F_\theta,\lambda_t$ & PF-ODE field $F_\theta(x,t)=f_t(x)-\tfrac12 g_t^2s_\theta(x,t)$ and half-log-SNR $\lambda_t=\log(\alpha_t/\sigma_t)$ for $\alpha_t,\sigma_t>0$.\\
$v_\theta,u_\theta$ & Instantaneous and interval-average velocity; flow-time direction and interval endpoints are specified in each table.\\
$G_\theta,H_\theta,\mathcal F_\theta$ & Generator or transition map, endpoint predictor, and consistency-model backbone, respectively.\\
\addlinespace[2pt]
$MSE,d_c$ & Mean squared error and pseudo-Huber discrepancy $d_c(x,y)=\sqrt{\|x-y\|_2^2+c^2}-c$, $c>0$.\\
$D_{\rm KL},\operatorname{MMD}^2$ & Kullback--Leibler divergence and squared maximum mean discrepancy between probability distributions.\\
$Adv(O_s,O_t)$ & Generator adversarial objective comparing generated outputs/features with real-data or teacher reference outputs/features.\\
$p_{\theta,t},q_t$ & Student and reference marginals under the same forward perturbation.\\
\addlinespace[2pt]
$\operatorname{Solver}_\phi,\operatorname{SDE}_\phi$ & Teacher ODE and SDE solvers; $\operatorname{MultiStepSDE}_\phi(x_t;t\to s,h)$ uses $h$ substeps.\\
$\operatorname{aDDIM}_{t\to s}$ & \begin{tabular}[t]{@{}l@{}}
$\hat\epsilon=(x_t-\alpha_t\hat x_0)/\sigma_t$, $b_{s,t}=\alpha_s-\alpha_t\sigma_s/\sigma_t$,\\
$\operatorname{aDDIM}_{t\to s}=\alpha_s\hat x_0+\sqrt{\sigma_s^2+d\,b_{s,t}^2V_t/\|\hat\epsilon\|_2^2}\,\hat\epsilon$.\\
$d$ is data dimension and $V_t$ is per-dimension clean-data uncertainty~\cite{heek2024multistep}.
\end{tabular}\\
$w,\omega$ & Time and discrepancy weights specified by each training objective.\\
$\mathcal D_t^v u,\mathrm{JVP}$ & Total derivative $(\partial_z u)v+\partial_tu$ at fixed interval start; evaluated by a Jacobian--vector product.\\
\addlinespace[2pt]
$\Phi_{s\to t}$ & Exact flow map, with $\Phi_{r\to t}\circ\Phi_{s\to r}=\Phi_{s\to t}$ and $\Phi_{t\to t}=\operatorname{Id}$.\\
$k,\varphi$ & Moment-matching kernel and learned feature map used by AYT, respectively.\\
$L_{\mathrm{TC}},L_{\mathrm{IQ}}$ & Text-consistency and image-quality losses in Q-Sched.\\
$\mathcal H_k,\mathrm{KV}$ & Generated history at video chunk $k$ and attention key/value states retained across chunks.\\
$\Delta x_{t\to r},\overline{F}_{t,r}$ & Finite state increment and interval-average ODE direction, $\Delta x_{t\to r}=(r-t)\overline{F}_{t,r}$.\\
$P_n,U_n,\bar G_n^{-1},\mathcal B$ & GeoRK2 feature projector, its basis, inverse metric, and feature update map, respectively.\\
$\mathcal T,\mathrm{NFE}$ & Sampling grid and number of function evaluations, respectively.\\
\bottomrule
\end{tabularx}
\label{symbols}
\end{table*}

\begin{enumerate}
\item \textbf{Distillation:} Uses a pretrained teacher to compress multi-step denoising into one or a few evaluations via step, consistency, adversarial, flow-map, or score distillation. Initializing the student from the teacher makes distillation far cheaper than training from scratch, while shifting the central challenge to distribution matching and low-step stability.

\item \textbf{Consistency Training:} Removes the two-stage pretraining-then-distillation pipeline by learning few-step or single-step transitions directly from data through self-consistency, interval-average velocity (mean-flow), or moment-matching objectives. These methods avoid holding a teacher model in memory during training and now scale to billion-parameter generators.

\item \textbf{Sampling Methods:} Accelerates a pretrained denoiser without retraining the full backbone by improving how the reverse ODE or SDE is discretized and executed. This category covers high-order numerical solvers, optimized timestep schedules, lightweight learned step corrections, and parallel trajectory execution.

\item \textbf{Denoising Network Design:} Reduces the compute of each individual denoiser evaluation or avoids redundant work across neighboring timesteps. Main approaches include linear and sparse attention mechanisms, adaptive feature caching and forecasting, low-bit quantization, and structured pruning.
\end{enumerate}

Section~\ref{sec:prelim} introduces the shared mathematical preliminaries across Markov, SDE/PF-ODE, consistency, rectified-flow, score-distillation, and adversarial formulations. Section~\ref{sec:methods} compares the four acceleration families in a shared notation. Section~\ref{sec:quantitative} compares published image and video benchmarks within their evaluation protocols. Sections~\ref{sec:future} and~\ref{sec:conclusion} discuss open directions and the main findings.
\section{Preliminaries}
\label{sec:prelim}
\subsection{Diffusion Models from Markov Perspective}
Denoising diffusion probabilistic models (DDPMs)~\cite{ho2020denoising} learn to generate samples by reversing a gradual noising process. DDPM defines a $T$-step forward process and a $T$-step reverse process. Its forward process perturbs the clean state \(x_{t_0}\) into a noisier state \(x_{t_n}\):
\begin{equation}
\begin{aligned}
q(x_{t_n}\mid x_{t_0})=\mathcal N(x_{t_n};\alpha_{t_n} x_{t_0},\sigma_{t_n}^2 I),\\
x_{t_n}={{\alpha}_{t_n}}x_{{t_0}}+\sigma_{t_n}\epsilon,\epsilon \sim \mathcal N(0,\rm \textbf{I}).
\end{aligned}
\end{equation}
For DDPM, define \(a_j=1-\beta_j\), \(\bar a_n=\prod_{j=1}^{n}a_j\), \(\alpha_{t_n}=\sqrt{\bar a_n}\), and \(\sigma_{t_n}=\sqrt{1-\bar a_n}\). These coefficients set the signal and noise scales at time \(t_n\).

The learned reverse process factors the joint density \(p_\theta(x_{0:T})\) into a prior and Gaussian transition kernels:
\begin{equation}
\begin{aligned} 
\label{eq:reverse}
 p_{\theta}({x}_{t_{n-1}}|{x}_{t_n}) &= \mathcal{N}({x}_{t_{n-1}}; \mu_{\theta}({x}_{t_n}, {t_n}), \Sigma_\theta({x}_{t_n}, {t_n})).
\end{aligned} 
\end{equation}

Here, the reverse variance can follow a fixed schedule such as \(\beta_{t_n}I\) or \(\tilde\beta_{t_n}I\), where \(\tilde\beta_{t_n}=\beta_n(1-\bar a_{n-1})/(1-\bar a_n)\) for adjacent original DDPM steps. Learned variances are also studied in~\cite{nichol2021improved}. A generative model with parameters \(\theta\) estimates the mean of the Gaussian distribution. For conditional generation, the condition \(c\) is supplied to the denoising network.

\subsection{Diffusion Models from a Stochastic Differential Equation Perspective}

In continuous time, the forward process follows a stochastic differential equation:
\begin{equation}
\mathrm d { {x}} = {f}_t({ {x}}) \, \mathrm d t + g_t \, \mathrm d\boldsymbol{w} \label{eq:sde-forward}
\end{equation}
where $f_{t}( {x}) = \frac{ \mathrm d \log \alpha_{t}}{ \mathrm d{t}} {x}$ and $g_{t}^2 = \frac{\mathrm d \sigma_t^2}{ \mathrm d t} - 2 \frac{ \mathrm d \log \alpha_t}{\mathrm d t} \sigma_{t}^2$.  
${\alpha}_t$ is the predefined scale factor and $\boldsymbol w_{t}$ denotes a standard Wiener process. $\sigma_{t}$ controls the level of noise.

The corresponding probability-flow ODE evolves the same time marginals under the exact score:
\begin{equation}
\mathrm d  {x} = \left( {f}_t( {x}) - \frac{1}{2} g_t^2 \nabla_{ {x}} \log p_t( {x}) \right) \, \mathrm d t \label{eq:flow-ode}
\end{equation}
Sampling integrates this probability-flow ODE (PF-ODE) from high to low diffusion time. For Gaussian perturbations, the noise predictor gives \(s_\theta(x_t,t)=-\epsilon_\theta(x_t,t)/\sigma_t\) for \(\sigma_t>0\)~\cite{song2021sde}. The corresponding clean-data prediction is \(D_\theta(x_t,t)=(x_t-\sigma_t\epsilon_\theta(x_t,t))/\alpha_t\) for \(\alpha_t>0\).

\subsection{Consistency Models}
Consistency models~\cite{song2023consistency} map states on a PF-ODE trajectory to its clean-side endpoint at a small positive time \(t_0\). At the endpoint, the map obeys
\begin{equation}
{f} : \left( {x}_{t_n}, {t_n}\right) \mapsto  {x}_{t_0}, \quad {t_n} \in [t_0, t_N].
\end{equation}

A skip connection enforces this boundary condition. With a neural backbone \(\mathcal F_\theta\), the map is
\begin{equation}
\label{eq:6}
f_\theta( {x}_{t_n}, {t_n}) = c_\text{skip}(t_n)  {x_{t_n}} + c_\text{out}(t_n) \mathcal{F}_\theta( {x_{t_n}}, t_n)
\end{equation}
where \( c_\text{skip}(t_n) \) and \( c_\text{out}(t_n) \) are differentiable functions such that \( c_\text{skip}(t_0) = 1 \), and \( c_\text{out}(t_0) = 0 \).

Consistency distillation matches predictions at neighboring points of a teacher trajectory. The target network follows an exponential moving average (EMA):

\begin{equation}
\resizebox{\linewidth}{!}{
$
\mathcal{L}_{\rm CD}=\mathbb{E}\left[w_n d\left(f_\theta(x_{t_{n+1}}, t_{n+1}),\operatorname{sg}\big(f_{\theta^-}(\hat x_{t_n}^{\phi},t_n)\big)\right)\right].
$
}
\end{equation}
Here \(w_n\) is the loss weight, \(d\) is a discrepancy, and \(\operatorname{sg}\) stops gradients. The frozen teacher's PF-ODE solver produces \(\hat x_{t_n}^{\phi}\) from \(x_{t_{n+1}}\), and \(\theta^-\) is the exponential moving average of \(\theta\).

\subsection{Rectified Flow} \label{sec2.3}

Rectified Flow~\cite{liu2022flow} formulates generative transport between a source distribution \(\pi_0\) and a target data distribution \(\pi_1\) through an ordinary differential equation (ODE). Given paired samples \((x_0, x_1) \sim \pi\) drawn from an initial coupling, intermediate states are constructed along straight segments \(x_t = (1 - t)x_0 + t x_1\) for \(t \in [0, 1]\). A parameterized velocity field \(v_\theta(x_t, t)\) is trained via least-squares regression to match the constant displacement vector \(x_1 - x_0\) of each linear path:
\begin{equation}
\label{eq:8}
\theta^* = \arg\min_\theta \mathbb{E}_{\substack{t \sim \mathcal{U}[0,1] \\ (x_0, x_1) \sim \pi}} \left[ \lVert v_\theta(x_t, t) - (x_1 - x_0) \rVert_2^2 \right].
\end{equation}
Wherever distinct linear paths intersect at \((x_t, t)\), the population minimizer resolves to the conditional expectation \(\mathbb{E}[x_1 - x_0 \mid x_t, t]\).

At inference time, samples are generated by evolving a prior draw \(x_0 \sim \pi_0\) from \(t = 0\) to \(t = 1\) along \(\mathrm{d}x_t / \mathrm{d}t = v_\theta(x_t, t)\), which is discretized with a numerical solver such as Euler's method:
\begin{equation}
x_{t_{n+1}} = x_{t_n} + (t_{n+1} - t_n) v_\theta(x_{t_n}, t_n).
\end{equation}
Because intersecting interpolants bend the learned ODE trajectories, recursively retraining the velocity field on synthetic endpoint pairs \((x_0, \hat{x}_1)\) simulated from the current ODE—a procedure known as \emph{reflow}—progressively straightens sampling trajectories and lowers convex transport costs without solving classical optimal transport~\cite{villani2009optimal} directly.

\subsection{Score Distillation}

Score distillation sampling was first introduced in DreamFusion~\cite{poole2022dreamfusion} to transfer the knowledge embedded in pretrained text-to-image diffusion models into a 3D Neural Radiance Field (NeRF). Later, the Diff-Instruct~\cite{luo2023diff} framework extended this idea by proposing a general method for score distillation. In this framework, a pretrained diffusion model is converted into a single-step generator by minimizing the Kullback–Leibler (KL) divergence between the teacher and the student distributions.

More specifically, Diff-Instruct introduces the \emph{Integral KL divergence} (IKL), which is defined as
\begin{equation}
\label{eq:9}
\resizebox{0.9\linewidth}{!}{$
\mathcal J(\theta)=\int_{t_0}^{t_N} w(t)\,D_{\mathrm{KL}}\big(p_{\theta,t}\,\|\,q_t\big)\,\mathrm dt
=\int_{t_0}^{t_N}w(t)\,\mathbb E_{x_t\sim p_{\theta,t}}\!\left[\log\frac{p_{\theta,t}(x_t)}{q_t(x_t)}\right]\,\mathrm dt,
$}
\end{equation}
where \(p_{\theta,t}\) and \(q_t\) denote the student and teacher marginal densities at time \(t\) under the same forward perturbation. The weighting function \(w(t)\) controls how the objective combines divergences across the diffusion time interval.

The gradient of the IKL with respect to the generator parameters $\theta$ can be expressed as
\begin{equation}
\label{eq:10}
\resizebox{0.9\linewidth}{!}{$\displaystyle\nabla_\theta\mathcal J=\int_{t_0}^{t_N} w(t)\,\mathbb E_{z,\epsilon}\!\left[(\partial_\theta x_t)^{\mathsf T}\big(s_{p_{\theta,t}}(x_t)-s_{q_t}(x_t)\big)\right]\,\mathrm dt.$}
\end{equation}
Here \(z\sim p_z\) is drawn from the prior, \(G_\theta(z)\) is a sample from the one-step student generator, and \(x_t=\alpha_t G_\theta(z)+\sigma_t\epsilon\), with \(\epsilon\sim\mathcal N(0,I)\). The functions \(s_{p_{\theta,t}}\) and \(s_{q_t}\) are the scores of the noised student and teacher distributions, respectively. An auxiliary model estimates the first score from perturbed student samples.

Training alternates between two updates. The first trains the auxiliary score model with a diffusion denoising loss on samples produced by the current student. The second uses the estimated score difference in Eq.~\eqref{eq:10} to update the student generator. The trained generator produces samples in one forward pass.

\subsection{Generative Adversarial Networks}
A generative adversarial network~\cite{goodfellow2020generative} trains a discriminator \(D\) to distinguish real samples from generated ones and a generator \(G\) to fool it. The minimax objective is
\[
\begin{aligned}\min_G\max_D\;&\mathbb E_{x\sim p_{\rm data}}[\log D(x)]\\&+\mathbb E_{z\sim p_z}[\log(1-D(G(z)))],\end{aligned}
\]
where \(z \sim \mathcal{N}(0, I)\) is drawn from a standard normal distribution.

When the real and generated distributions have disjoint support, the discriminator can separate them without providing a useful training signal. DiffusionGAN~\cite{wang2022diffusion} perturbs both sets of samples with the same forward diffusion process before discrimination.

\newcommand{\ImageComparisonTable}{%
\begin{table*}[!tp]
\centering
\small
\renewcommand{\arraystretch}{1.02}
\setlength{\tabcolsep}{3pt}
\caption{Quantitative comparison of diffusion acceleration. Panels (a)--(c) cover image generation and sampler schedules. Arrows indicate the direction of improvement.}
\label{tab:quantitative}
\begin{minipage}[t]{0.487\linewidth}
\setlength{\tabcolsep}{2pt}
\comparisonpanel{(a) One-evaluation ImageNet $256\times256$\\Geng et al.~\cite{geng2026improvedmeanflow}, Table~2}
\begin{tabular*}{\linewidth}{@{\extracolsep{\fill}}lrrrr@{}}
\toprule
\textbf{Configuration} & \shortstack[r]{\textbf{Params}\\\textbf{(M)}} & \textbf{GFLOPs} & \textbf{FID$\downarrow$} & \textbf{IS$\uparrow$} \\
\midrule
MF-B/2 & 131 & 23.1 & 6.17 & 208.0 \\
MF-M/2 & 308 & 54.0 & 5.01 & 252.0 \\
MF-L/2 & 459 & 80.9 & 3.84 & 250.9 \\
MF-XL/2 & 676 & 119.0 & 3.43 & 247.5 \\
iMF-B/2 & 89 & 24.9 & 3.39 & 255.3 \\
iMF-M/2 & 174 & 49.9 & 2.27 & 257.7 \\
iMF-L/2 & 409 & 116.4 & 1.86 & 276.6 \\
iMF-XL/2 & 610 & 174.6 & 1.72 & 282.0 \\
\bottomrule
\end{tabular*}
\end{minipage}\hfill%
\begin{minipage}[t]{0.487\linewidth}
\setlength{\tabcolsep}{2pt}
\comparisonpanel{(b) SDXL / 10K COCO 2014 prompts\\Yin et al.~\cite{yin2024improved}, Table~2}
\begin{tabular*}{\linewidth}{@{\extracolsep{\fill}}lrrrr@{}}
\toprule
\textbf{Method} & \shortstack[r]{\textbf{Forward}\\\textbf{passes}} & \textbf{FID$\downarrow$} & \textbf{Patch FID$\downarrow$} & \textbf{CLIP$\uparrow$} \\
\midrule
LCM-SDXL & 1 & 81.62 & 154.40 & 0.275 \\
LCM-SDXL & 4 & 22.16 & 33.92 & 0.317 \\
SDXL-Turbo & 1 & 24.57 & 23.94 & 0.337 \\
SDXL-Turbo & 4 & 23.19 & 23.27 & 0.334 \\
SDXL-Lightning & 1 & 23.92 & 31.65 & 0.316 \\
SDXL-Lightning & 4 & 24.46 & 24.56 & 0.323 \\
DMD2 & 1 & 19.01 & 26.98 & 0.336 \\
DMD2 & 4 & 19.32 & 20.86 & 0.332 \\
SDXL, CFG=6 & 100 & 19.36 & 21.38 & 0.332 \\
SDXL, CFG=8 & 100 & 20.39 & 23.21 & 0.335 \\
\bottomrule
\end{tabular*}
\end{minipage}

\comparisonpanel{(c) ImageNet $256\times256$, CFG=2.0: Sabour et al.~\cite{sabour2024ays}, Table~1}
{\setlength{\tabcolsep}{2.3pt}
\begin{tabular*}{\textwidth}{@{\extracolsep{\fill}}llrrrrrrrrr@{}}
\toprule
\textbf{Sampler} & \textbf{Schedule} & \multicolumn{3}{c}{\textbf{NFE=10}} & \multicolumn{3}{c}{\textbf{NFE=20}} & \multicolumn{3}{c}{\textbf{NFE=30}} \\
\cmidrule(lr){3-5}\cmidrule(lr){6-8}\cmidrule(lr){9-11}
 &  & FID$\downarrow$ & sFID$\downarrow$ & IS$\uparrow$ & FID$\downarrow$ & sFID$\downarrow$ & IS$\uparrow$ & FID$\downarrow$ & sFID$\downarrow$ & IS$\uparrow$ \\
\midrule
Stochastic DDIM & EDM & 66.71 & 126.92 & 25.04 & 17.42 & 49.89 & 152.74 & 9.85 & 26.15 & 242.81 \\
 & Uniform & 24.48 & 67.96 & 112.53 & 9.32 & 22.65 & 256.27 & 8.41 & 13.67 & 299.44 \\
 & AYS & 23.13 & 64.37 & 118.61 & 8.96 & 19.78 & 264.98 & 8.29 & 11.65 & 304.37 \\
\addlinespace[2pt]
SDE-DPM++(2M) & EDM & 8.48 & 21.83 & 214.49 & 7.05 & 8.17 & 307.41 & 7.55 & 6.58 & 325.78 \\
 & Uniform & 8.47 & 13.36 & 243.09 & 7.63 & 11.02 & 282.77 & 7.14 & 8.59 & 305.57 \\
 & AYS & 6.11 & 8.48 & 281.44 & 6.79 & 5.93 & 322.92 & 7.28 & 5.48 & 330.01 \\
\addlinespace[2pt]
ER-SDE-Solver 1 & EDM & 17.78 & 35.25 & 147.57 & 6.99 & 12.70 & 255.69 & 6.20 & 8.51 & 282.52 \\
 & Uniform & 8.79 & 18.33 & 222.93 & 6.25 & 8.19 & 280.74 & 6.09 & 6.56 & 293.47 \\
 & AYS & 8.36 & 15.91 & 266.44 & 6.06 & 7.28 & 282.06 & 5.87 & 5.97 & 295.40 \\
\addlinespace[2pt]
ER-SDE-Solver 2 & EDM & 7.36 & 14.19 & 231.46 & 5.58 & 6.33 & 290.80 & 5.85 & 5.69 & 299.12 \\
 & Uniform & 5.28 & 6.19 & 277.57 & 5.56 & 5.55 & 295.69 & 5.72 & 5.50 & 300.25 \\
 & AYS & 5.38 & 6.24 & 275.35 & 5.45 & 5.19 & 297.78 & 5.71 & 5.16 & 301.79 \\
\addlinespace[2pt]
ER-SDE-Solver 3 & EDM & 6.94 & 13.01 & 237.70 & 5.58 & 6.13 & 292.75 & 5.87 & 5.61 & 299.33 \\
 & Uniform & 5.13 & 6.08 & 277.65 & 5.52 & 5.57 & 295.94 & 5.71 & 5.48 & 301.52 \\
 & AYS & 5.28 & 6.10 & 275.80 & 5.47 & 5.17 & 298.05 & 5.73 & 5.14 & 302.40 \\
\addlinespace[2pt]
DDIM (ODE) & Uniform & 7.57 & 14.53 & 224.50 & 5.39 & 7.08 & 273.33 & 5.23 & 5.87 & 283.27 \\
 & AYS & 6.96 & 12.21 & 226.25 & 5.09 & 12.21$^\dagger$ & 273.94 & 4.99 & 5.53 & 283.37 \\
\addlinespace[2pt]
DPM++(2M) & LogSNR & 4.82 & 6.83 & 252.71 & 4.81 & 5.41 & 287.20 & 4.98 & 5.22 & 288.81 \\
 & AYS & 4.31 & 6.64 & 260.32 & 4.70 & 5.34 & 284.17 & 4.96 & 5.15 & 290.65 \\
\bottomrule
\end{tabular*}
\par\smallskip
\begin{minipage}{\linewidth}\footnotesize
$^\dagger$ Sabour et al., Table~1, lists 12.21 at both 10 and 20 NFE for DDIM (ODE) with AYS.
\end{minipage}
}
\end{table*}
}

\newcommand{\VideoComparisonTable}{%
\begin{table*}[!tp]
\ContinuedFloat
\centering
\small
\renewcommand{\arraystretch}{1.02}
\setlength{\tabcolsep}{3pt}
\caption[]{Quantitative comparison of diffusion acceleration (continued): video caching and generation.}
\label{tab:quantitative-video}
\comparisonpanel{(d) Video caching, 50-step schedules: Haghighi and Alahi~\cite{haghighi2026sencache}, Table~1}
\begin{tabular*}{\textwidth}{@{\extracolsep{\fill}}lrrrrr@{}}
\toprule
\textbf{Method} & \textbf{NFE$\downarrow$} & \textbf{Cache ratio} & \textbf{LPIPS$\downarrow$} & \textbf{PSNR$\uparrow$} & \textbf{SSIM$\uparrow$} \\
\midrule
\rowcolor{black!9}
\multicolumn{6}{l}{Wan 2.1: 81 frames, $832\times480$}\\
TeaCache-slow & 33 & 34\% & 0.0384 & 30.6097 & 0.9372 \\
MagCache-slow & 25 & 50\% & 0.0390 & 30.7226 & 0.9387 \\
SenCache-slow & 25 & 50\% & 0.0390 & 30.7233 & 0.9387 \\
TeaCache-fast & 25 & 50\% & 0.0966 & 25.0661 & 0.8697 \\
MagCache-fast & 21 & 58\% & 0.0603 & 28.3684 & 0.9143 \\
SenCache-fast & 21 & 58\% & 0.0540 & 29.1400 & 0.9219 \\
\midrule
\rowcolor{black!9}
\multicolumn{6}{l}{CogVideoX: 49 frames, $720\times480$}\\
TeaCache & 22 & 56\% & 0.5855 & 14.0194 & 0.5702 \\
MagCache & 23 & 54\% & 0.1952 & 21.8502 & 0.7332 \\
SenCache & 22 & 56\% & 0.1901 & 22.0900 & 0.7786 \\
\midrule
\rowcolor{black!9}
\multicolumn{6}{l}{LTX-Video: 161 frames, $768\times512$}\\
TeaCache & 32 & 36\% & 0.2763 & 20.2019 & 0.7270 \\
MagCache & 28 & 44\% & 0.1795 & 23.3655 & 0.8224 \\
SenCache & 27 & 46\% & 0.1625 & 23.6660 & 0.8293 \\
\bottomrule
\end{tabular*}
\comparisonpanel{(e) Short video / VBench, one H100: Huang et al.~\cite{huang2025selfforcing}, Table~1}
\begin{tabular*}{\textwidth}{@{\extracolsep{\fill}}lrrrrrrr@{}}
\toprule
\textbf{Method} & \textbf{Params (B)} & \textbf{Resolution} & \textbf{FPS$\uparrow$} & \textbf{First-frame latency (s)$\downarrow$} & \textbf{Total$\uparrow$} & \textbf{Quality$\uparrow$} & \textbf{Semantic$\uparrow$} \\
\midrule
LTX-Video & 1.9 & $768\times512$ & 8.98 & 13.5 & 80.00 & 82.30 & 70.79 \\
Wan 2.1 & 1.3 & $832\times480$ & 0.78 & 103 & 84.26 & 85.30 & 80.09 \\
SkyReels-V2 & 1.3 & $960\times540$ & 0.49 & 112 & 82.67 & 84.70 & 74.53 \\
MAGI-1 & 4.5 & $832\times480$ & 0.19 & 282 & 79.18 & 82.04 & 67.74 \\
CausVid (Wan 1.3B) & 1.3 & $832\times480$ & 17.0 & 0.69 & 81.20 & 84.05 & 69.80 \\
Self Forcing, chunk-wise & 1.3 & $832\times480$ & 17.0 & 0.69 & 84.31 & 85.07 & 81.28 \\
NOVA & 0.6 & $768\times480$ & 0.88 & 4.1 & 80.12 & 80.39 & 79.05 \\
Pyramid Flow & 2.0 & $640\times384$ & 6.7 & 2.5 & 81.72 & 84.74 & 69.62 \\
Self Forcing, frame-wise & 1.3 & $832\times480$ & 8.9 & 0.45 & 84.26 & 85.25 & 80.30 \\
\bottomrule
\end{tabular*}
\comparisonpanel{(f) 30-second video / VBench-Long, one H100: Yang et al.~\cite{yang2026longlive}, Table~3}
\begin{tabular*}{\textwidth}{@{\extracolsep{\fill}}lrrrr@{}}
\toprule
\textbf{Method} & \textbf{FPS$\uparrow$} & \textbf{Total$\uparrow$} & \textbf{Quality$\uparrow$} & \textbf{Semantic$\uparrow$} \\
\midrule
SkyReels-V2 & 0.49 & 75.29 & 80.77 & 53.37 \\
FramePack & 0.92 & 81.95 & 83.61 & 75.32 \\
Self Forcing & 17.0 & 81.59 & 83.82 & 72.70 \\
LongLive & 20.7 & 83.52 & 85.44 & 75.82 \\
\bottomrule
\end{tabular*}
\end{table*}
}

\section{Methodology}
\label{sec:methods}
\subsection{Distillation}
\begin{table*}[!b]
\centering
\caption{Distillation Methods, Part A. Consistency and flow distillation.}
\label{tab:distilloverview}
\setlength{\tabcolsep}{3pt}
\renewcommand{\arraystretch}{1.04}
\resizebox{\linewidth}{!}{%
\begin{tabular}{@{}M{100pt} M{187pt} M{111pt} M{127pt} M{163pt}@{}}
\toprule
\textbf{Model} & \textbf{Preliminaries} & \textbf{Student Output}~($O_s$) & \textbf{Teacher Output}~($O_t$) & \textbf{Loss Function} \\
\midrule
\rowcolor{black!9}
\multicolumn{5}{@{}l}{\hspace{3pt}\textbf{Consistency Distillation}}\\
CM~\cite{song2023consistency} & $\hat x^\phi_{t_n}=x_{t_{n+1}}+\Delta t\,V_{n+1}^\phi$ & $f_\theta(x_{t_{n+1}},t_{n+1})$ & $f_{\theta^-}(\hat x^\phi_{t_n},t_n)$ & $\lambda^{CM}(t_n)MSE(O_s,O_t)$ \\
CD (pseudo-Huber)~\cite{song2023improved} & $\hat x^\phi_{t_n}=x_{t_{n+1}}+\Delta t\,V_{n+1}^\phi$ & $f_\theta(x_{t_{n+1}},t_{n+1})$ & $f_{\theta^-}(\hat x^\phi_{t_n},t_n)$ & $\lambda^{\mathrm{iCT}}(t_n)d_c(O_s,O_t)$ \\
SCott~\cite{liu2024scott} & $\hat x^\phi_{t_m}=MultiStepSDE(x_{t_n},t_n,t_m,h;\Phi)$ & $f_\theta(x_{t_n},t_n)$ & $f_{\theta^-}(\hat x^\phi_{t_m},t_m)$ & \begin{tabular}[c]{@{}l@{}}$\lambda^{CM}(t_n)MSE(O_s,O_t)$\\$+\lambda^{SCott}Adv(O_s,x_0)$\end{tabular} \\
CFM~\cite{yang2024consistency} & \begin{tabular}[c]{@{}l@{}}$\hat x^\phi_{t_n}=x_{t_{n+1}}+\Delta t\,V_{n+1}^\phi$\\$f_\theta(x,t)=x-tv_\theta(x,t),\quad v_\theta^{t}=v_\theta(x_t,t)$\end{tabular} & $f_\theta(x_{t_{n+1}},t_{n+1})$ & $f_{\theta^-}(\hat x^\phi_{t_n},t_n)$ & \begin{tabular}[c]{@{}l@{}}$MSE(O_s,O_t)$\\$+\lambda^{CFM}MSE(v_\theta^{t_{n+1}},v_{\theta^-}(\hat x^\phi_{t_n},t_n))$\end{tabular} \\
MCM~\cite{heek2024multistep} & \begin{tabular}[c]{@{}l@{}}$\hat x^\phi_{t_n}=aDDIM(x_{t_{n+1}},t_{n+1},t_n;\Phi)$\\$\hat x_{ref}=f_{SG(\theta)}(\hat x^\phi_{t_n},t_n)$\\$c=\sigma_{t^k_{step}}/\sigma_{t_{n+1}}$\end{tabular} & $f_\theta(x_{t_{n+1}},t_{n+1})$ & \begin{tabular}[c]{@{}l@{}}$\alpha_{t^k_{step}}\hat x_{ref}$\\$\quad +\frac{\sigma_{t^k_{step}}}{\sigma_{t_n}}(\hat x^\phi_{t_n}-\alpha_{t_n}\hat x_{ref})$\\$\qquad(t_n>t^k_{step})$\\$O_t=\hat x^\phi_{t_n},\quad t_n=t^k_{step}$\end{tabular} & \begin{tabular}[c]{@{}l@{}}$\lambda^{MCM}(t_n)d_{c_H}\!\bigl(O_s,$\\$\quad SG\!\left(\frac{O_t-cx_{t_{n+1}}}{\alpha_{t^k_{step}}-c\alpha_{t_{n+1}}}\right)\bigr)$\end{tabular} \\
CTM~\cite{kim2023consistency} & \begin{tabular}[c]{@{}l@{}}$\hat x^\phi_{t_n}=CTMSolver($\\$\qquad x_{t_{n+m}},t_{n+m},t_n;\phi)$\\$s\leq t_n$\end{tabular} & \begin{tabular}[c]{@{}l@{}}$f_{SG(\theta)}(u_s,s,0)$\\$u_s=f_\theta(x_{t_{n+m}},t_{n+m},s)$\end{tabular} & \begin{tabular}[c]{@{}l@{}}$f_{SG(\theta)}(\hat u_s,s,0)$\\$\hat u_s=f_{SG(\theta)}(\hat x^\phi_{t_n},t_n,s)$\end{tabular} & \begin{tabular}[c]{@{}l@{}}$d(O_s,O_t)+\lambda_1^{CTM}Adv(O_s,x_0)$\\$+\lambda_2^{CTM}L_{DSM}$\end{tabular} \\
Hyper-SD~\cite{ren2024hyper} & \begin{tabular}[c]{@{}l@{}}$\hat x^\phi_{t_n}=Solver(x_{t_{n+1}},t_{n+1},t_n;\Phi)$\\$f_{\theta,s}(x,t)=DDIM(x,t,s;\theta)$\\$b=t^k_{step}\leq s\leq t_n$\end{tabular} & $f_{\theta,s}(x_{t_{n+1}},t_{n+1})$ & $f_{\theta^-,s}(\hat x^\phi_{t_n},t_n)$ & \begin{tabular}[c]{@{}l@{}}$L_{TSCD}=d(O_s,O_t)$\\$L_{feedback}=[a-R(\hat x_0,c)]_++L_{percep}$\\$L_{DMD}=MSE(\hat x_0,SG(\hat x_0-\Delta_{DM}))$\end{tabular} \\
PCM~\cite{wang2024phased} & \begin{tabular}[c]{@{}l@{}}$\hat x^\phi_{t_n}=x_{t_{n+1}}+\Delta t\,V_{n+1}^\phi$\\$b=t^k_{step}$\\$\tilde O_s=Q_{b\to u}(O_s),\quad\tilde O_t=Q_{b\to u}(O_t)$\end{tabular} & $f_{\theta,b}(x_{t_{n+1}},t_{n+1})$ & $f_{\theta^-,b}(\hat x^\phi_{t_n},t_n)$ & $MSE(O_s,O_t)+\lambda^{PCM}Adv(\tilde O_s,\tilde O_t)$ \\
CMT~\cite{hu2026cmt} & \begin{tabular}[c]{@{}l@{}}$\hat x^\Phi_{t_n}=Solver(x_{t_N},t_N,t_n;\Phi),$\\$\hat x^\Phi_{t_0}=Solver(x_{t_N},t_N,t_0;\Phi)$\end{tabular} & $f_\theta(\hat x^\Phi_{t_n},t_n)$ & $\hat x^\Phi_{t_0}$ & $d(O_s,O_t)$ \\
\midrule
\rowcolor{black!9}
\multicolumn{5}{@{}l}{\hspace{3pt}\textbf{Flow Models}}\\
SlimFlow~\cite{zhu2025slimflow} & \begin{tabular}[c]{@{}l@{}}$\Phi\leftarrow AnnealingReflow$\\$\hat x^\Phi_{t_0}=ODE[v_\Phi](x_{t_N})$\\$x_{t_n}=x_{t_N}-(1-t_n)v_\Phi(x_{t_N},t_N)$\\$t_N=1$\end{tabular} & $x_{t_N}-v_\theta(x_{t_N},t_N)$ & $x_{t_n}-t_nv_\Phi(x_{t_n},t_n)$ & $L_{dist}+\mathbb D(O_s,O_t)$ \\
InstaFlow~\cite{liu2023instaflow} & \begin{tabular}[c]{@{}l@{}}$\hat x_1=x_0+\int_0^1v_\Phi(x_t,t\mid\mathcal T)\,dt,$\\$ x_{t_n}=(1-t_n)x_0+t_n\hat x_1$\end{tabular} & $x_0+v_\theta(x_0,0\mid\mathcal T)$ & $\hat x_1$ & $MSE(O_s,O_t)$ \\
PeRFlow~\cite{yan2024perflow} & \begin{tabular}[c]{@{}l@{}}$x^\Phi_{t_{n-1}}=Solver(x_{t_n},t_n,t_{n-1};\Phi)$\\$x_t=x_{t_n}+\frac{t-t_n}{t_{n-1}-t_n}(x^\Phi_{t_{n-1}}-x_{t_n})$\end{tabular} & $v_\theta(x_t,t)$ & $\frac{x^\Phi_{t_{n-1}}-x_{t_n}}{t_{n-1}-t_n}$ & $MSE(O_s,O_t)$ \\
FreeFlow~\cite{tong2026flowmap} & \begin{tabular}[c]{@{}l@{}}$G_\theta(z,\delta)=z+\delta A_\theta(z,\delta),\quad g_\Phi=-F_\Phi$\\$y_r=I_r(G_\theta(z,1),\epsilon)$\end{tabular} & $A_\theta(z,\delta)$ & \begin{tabular}[c]{@{}l@{}}$g_\Phi(G_\theta(z,\delta),1-\delta)$\\$-\delta\partial_\delta A_\theta(z,\delta)$\end{tabular} & \begin{tabular}[c]{@{}l@{}}$MSE(O_s,SG(O_t))$\\$+\lambda A_\theta(z,1)^T SG(d_r)$\\$d_r=g_{\theta^{fake}}(y_r,r)-g_\Phi(y_r,r)$\end{tabular} \\
MeanFlow + RAE~\cite{hu2026mfrae} & \begin{tabular}[c]{@{}l@{}}$z_0=E_{RAE}(x_0)$\\$z_{t_n}=(1-t_n)z_0+t_n\epsilon$\\$w=v_\Phi(z_{t_n},t_n)$\\$u^\pm=u_{\theta^-}(z_{t_n}\pm hw,r,t_n\pm h)$\\$\delta_hu=(u^+-u^-)/(2h)$\end{tabular} & $u_\theta(z_{t_n},r,t_n)$ & $w-(t_n-r)\delta_hu$ & $MSE(O_s,SG(O_t))$ \\
\bottomrule
\end{tabular}}
\par\smallskip
\begin{minipage}{\linewidth}\footnotesize
For the Euler rows, $\Delta t=t_n-t_{n+1}$ and $V_{n+1}^\phi=\Phi(x_{t_{n+1}},t_{n+1};\phi)$. $MSE$, $Adv$, and $SG$ denote squared-error regression, a generator adversarial objective, and stop-gradient. $O_t$ denotes teacher outputs or reference data and features. MCM uses the smooth Huber distance $d_{c_H}$ with $c_H>0$. CTM uses an EMA target in $f_{SG(\theta)}$, with $L_{DSM}=MSE(D_\theta(x_t,t),x_0)$ and feature distance $d$. PCM uses $Q_{b\to u}(x)=\alpha_u x/\alpha_b+\sqrt{\sigma_u^2-\alpha_u^2\sigma_b^2/\alpha_b^2}\epsilon$. Hyper-SD lists successive objectives, with $L_{percep}=\mathbb E[L_{instance}(m_I(\hat x_0),GT(x_0))]$ where $x_0$ is a real image and $\hat x_0$ is the generator output from its noisy input $q_{t_a}(\cdot\mid x_0)$. SlimFlow uses $L_{dist}=\mathbb D\bigl(ODE[v_\Phi](x_{t_N}),O_s\bigr)$. CFM uses reversed flow time, and InstaFlow uses noise-to-data time.
\end{minipage}
\end{table*}

\begin{table*}[!tp]
\ContinuedFloat
\centering
\caption{Distillation Methods, Part B (continued). Step, score, and adversarial distillation.}
\label{tab:distilloverview-b}
\setlength{\tabcolsep}{3pt}
\renewcommand{\arraystretch}{0.98}
\resizebox{\linewidth}{!}{%
\begin{tabular}{@{}M{100pt} M{187pt} M{111pt} M{127pt} M{163pt}@{}}
\toprule
\textbf{Model} & \textbf{Preliminaries} & \textbf{Student Output}~($O_s$) & \textbf{Teacher Output}~($O_t$) & \textbf{Loss Function} \\
\midrule
\rowcolor{black!9}
\multicolumn{5}{@{}l}{\hspace{3pt}\textbf{Step Distillation}}\\
PD~\cite{salimans2022progressive} & \begin{tabular}[c]{@{}l@{}}$\Phi^{(i+1)}\leftarrow\theta^{(i)},\quad\theta^{(i+1)}\leftarrow\Phi^{(i+1)}$\\$b=t_{n-2m},\quad c_{b,n}=\alpha_b-\alpha_{t_n}\sigma_b/\sigma_{t_n}$\\$\lambda^{PD}=w_{clean}/c_{b,n}^2$\end{tabular} & $DDIM(x_{t_n},t_n,t_{n-2m};\theta)$ & \begin{tabular}[c]{@{}l@{}}$DDIM(y,t_{n-m},t_{n-2m};\Phi)$\\$y=DDIM(x_{t_n},t_n,t_{n-m};\Phi)$\end{tabular} & $\lambda^{PD}(t_n)MSE(O_s,O_t)$ \\
SMD~\cite{zhang2026onpolicy} & \begin{tabular}[c]{@{}l@{}}$x_{t_n}=SG(Rollout_\theta(x_{t_N},t_n)),$\\$0\leq\lambda^{SMD}<1$\\$\Delta=t_n-t_{n-m},\quad a_\psi(\tau)=\sum_j a_jb_j(\tau)$\\$\int_0^1b_j(\tau)d\tau=0$\\$x_\tau=x_{t_n}-\Delta\int_0^\tau[v_\theta(x_{t_n},t_n)+a_\psi(q)]dq$\end{tabular} & $v_\theta(x_{t_n},t_n)+a_\psi(\tau)$ & $v_\Phi(x_\tau,t_n-\tau\Delta)$ & \begin{tabular}[c]{@{}l@{}}$MSE(O_s,O_t)$\\$-\lambda^{SMD}MSE(O_s,\tilde v_\Phi(x_\tau,t_n-\tau\Delta))$\end{tabular} \\
\midrule
\rowcolor{black!9}
\multicolumn{5}{@{}l}{\hspace{3pt}\textbf{Score Distillation}}\\
SwiftBrush~\cite{nguyen2024swiftbrush} & \begin{tabular}[c]{@{}l@{}}$\hat x^\theta_{t_0}=G_\theta(z),$\\$\hat x^\theta_{t_n}=\alpha_{t_n}\hat x^\theta_{t_0}+\sigma_{t_n}\epsilon$\end{tabular} & $w_{\rm VSD}(t_n)(\epsilon_{\Phi,n}-\epsilon_{\theta^{fake},n})$ & $\hat x^\theta_{t_0}$ & $MSE(O_t,SG(O_t-O_s))$ \\
DMD~\cite{yin2024one} & \begin{tabular}[c]{@{}l@{}}$\hat x^\theta_{t_0}=G_\theta(z),$\\$\hat x^\theta_{t_n}=\alpha_{t_n}\hat x^\theta_{t_0}+\sigma_{t_n}\epsilon$\end{tabular} & $\alpha_{t_n}(f_{\theta^{fake},n}-f_{\Phi,n})/w$ & $\hat x^\theta_{t_0}$ & \begin{tabular}[c]{@{}l@{}}$MSE(O_t,SG(O_t-O_s))$\\$+\lambda_{reg}d_{LPIPS}(O_t,\hat x^\Phi_{t_0}(z))$\end{tabular} \\
DMD2~\cite{yin2024improved} & \begin{tabular}[c]{@{}l@{}}$\hat x^\theta_{t_0}=G_\theta(z),$\\$\hat x^\theta_{t_n}=\alpha_{t_n}\hat x^\theta_{t_0}+\sigma_{t_n}\epsilon$\end{tabular} & $(f_{\theta^{fake},n}-f_{\Phi,n})/w$ & $\hat x^\theta_{t_0}$ & \begin{tabular}[c]{@{}l@{}}$MSE(O_t,SG(O_t-O_s))$\\$+\lambda^{DMD2}Adv(O_t,x_0)$\end{tabular} \\
SiD~\cite{zhou2024score} & \begin{tabular}[c]{@{}l@{}}$\hat x^\theta_{t_0}=G_\theta(z)$\\$\hat x^\theta_{t_n}=\hat x^\theta_{t_0}+\sigma_{t_n}\epsilon$\end{tabular} & $f_{\Phi,n}-f_{\theta^{fake},n}$ & $\hat x^\theta_{t_0}$ & \begin{tabular}[c]{@{}l@{}}$\lambda^{SiD}_1\|O_s\|_2^2$\\$+\lambda^{SiD}_2O_s^T(f_{\theta^{fake},n}-O_t)$\end{tabular} \\
eDMD~\cite{zhao2026edmd} & \begin{tabular}[c]{@{}l@{}}$\hat x^{\theta,k}_{t_0}=f_\theta(SG(x_{t_k}),t_k),\quad t_k\to t_{k-1}$\\$y_\tau=(1-\tau)\hat x^{\theta,k}_{t_0}+\tau\epsilon$\\$0<\tau<1,\quad\tau\geq t_{k-1}$\\$y_\tau(x)=(1-\tau)x+\tau\epsilon$\\$\min_{\theta^{fake}}\mathbb E_{x\sim p_{\theta,0}}\bigl[$\\$\qquad \lambda(\tau)MSE(f_{\theta^{fake}}(y_\tau(x),\tau),x)\bigr]$\end{tabular} & \begin{tabular}[c]{@{}l@{}}$(f_{\theta^{fake}}-f^{PT}_\Phi)/w_k$\\$\text{evaluated at }(y_\tau,\tau)$\end{tabular} & $\hat x^{\theta,k}_{t_0}$ & $MSE(O_t,SG(O_t-O_s))$ \\
VA-TDM~\cite{luo2026vatdm} & \begin{tabular}[c]{@{}l@{}}$x_r=DDIM(x_{t_n},t_n,r;\theta),$\\$y_\tau=a_{\tau|r}x_r+b_{\tau|r}\epsilon$\\$\hat x_0=f_\theta(x_{t_n},t_n),\quad0\leq r<\tau<1$\end{tabular} & $\{v_\tau^F-\lambda\bar v_{\theta,\tau}^F\}_{F=A,B}$ & \begin{tabular}[c]{@{}l@{}}$\{(1+\lambda_r)d^F_\tau\}_{F=A,B}$\\$d^F_\tau=v_\Phi(y_\tau,\tau)-h_{\theta^{fake}}^F(y_\tau)$\end{tabular} & $-\sum_{F=A,B}\pi_F SG(O_t^F)^T O_s^F$ \\
CausVid~\cite{yin2025causvid} & \begin{tabular}[c]{@{}l@{}}$x_0\sim p_{data},\quad x_{t_i}^i=\alpha_{t_i}x_0^i+\sigma_{t_i}\epsilon_i$\\$\hat x^\theta_{t_0}=G_\theta^{causal}(\{x_{t_i}^i\},\{t_i\})$\\$\hat x^\theta_{t_n}=\alpha_{t_n}\hat x^\theta_{t_0}+\sigma_{t_n}\epsilon$\end{tabular} & $\frac{\alpha_{t_n}^2}{\sigma_{t_n}^2}(f_{\theta^{fake},n}-f_{\Phi,n})$ & $\hat x^\theta_{t_0}$ & $MSE(O_t,SG(O_t-O_s))$ \\
SwiftBrush v2~\cite{dao2025swiftbrush} & \begin{tabular}[c]{@{}l@{}}$\hat x^\theta_{t_0}=G_\theta(z),$\\$\hat x^\theta_{t_n}=\alpha_{t_n}\hat x^\theta_{t_0}+\sigma_{t_n}\epsilon$\end{tabular} & $w_{\rm VSD}(t_n)(\epsilon_{\Phi,n}-\epsilon_{\theta^{fake},n})$ & $\hat x^\theta_{t_0}$ & \begin{tabular}[c]{@{}l@{}}$MSE(O_t,SG(O_t-O_s))$\\$+\lambda^{CLIP}[\tau_c-C(O_t,c)]_+$\end{tabular} \\
SiDA~\cite{zhou2024adversarial} & \begin{tabular}[c]{@{}l@{}}$\hat x^\theta_{t_0}=G_\theta(z),\quad x_t(y)=\alpha_ty+\sigma_t\epsilon$\\$\hat x^\theta_{t_n}=x_{t_n}(\hat x^\theta_{t_0})$\\$q_j=\omega(t_j)\alpha_{t_j}^2/(2\sigma_{t_j}^4)$\\$\bar q_B=|B|^{-1}\sum_{j\in B}q_j$\end{tabular} & $f_{\Phi,n}-f_{\theta^{fake},n}$ & $\hat x^\theta_{t_0}$ & \begin{tabular}[c]{@{}l@{}}$\displaystyle\frac{1}{|B|}\sum_{j\in B}\bigl[\lambda^{SiDA}_{1,j}\|O_{s,j}\|_2^2$\\$\qquad+\lambda^{SiDA}_{2,j}O_{s,j}^T(f_{\theta^{fake},j}-O_{t,j})\bigr]$\\$+\lambda^{adv}\bar q_B Adv_B(\{\hat x^\theta_{t_j}\})$\end{tabular} \\
SIM~\cite{luo2025one} & \begin{tabular}[c]{@{}l@{}}$\hat x^\theta_{t_0}=G_\theta(z),$\\$\hat x^\theta_{t_n}=\alpha_{t_n}\hat x^\theta_{t_0}+\sigma_{t_n}\epsilon$\end{tabular} & \begin{tabular}[c]{@{}l@{}}$s_{\theta^{fake}}(\hat x^\theta_{t_n},t_n)$\\$-s_\Phi(\hat x^\theta_{t_n},t_n)$\end{tabular} & \begin{tabular}[c]{@{}l@{}}$s_{\theta^{fake}}(\hat x^\theta_{t_n},t_n)$\\$+\epsilon/\sigma_{t_n}$\end{tabular} & $-\lambda(t_n)\frac{O_s^TO_t}{\sqrt{\|O_s\|_2^2+c^2}}$ \\
VarFlow~\cite{shao2025varflow} & $X,X^\prime\stackrel{iid}{\sim}p_{\theta,t_n},\quad Y\sim q_{t_n},\quad0<\beta<2$ & $(X,X^\prime)$ & $Y$ & $\lambda(t_n)(\|X-Y\|_2^\beta-\tfrac12\|X-X^\prime\|_2^\beta)$ \\
Self Forcing~\cite{huang2025selfforcing} & \begin{tabular}[c]{@{}l@{}}$\hat x^{\theta,i}_{t_0}=G_\theta(z_i;SG(\hat x^{\theta,<i}_{t_0}),c)$\\$p_\theta=\prod_i p_\theta(\hat x^{\theta,i}_{t_0}\mid\hat x^{\theta,<i}_{t_0},c)$\\$\hat x^\theta_{t_n}=\alpha_{t_n}\hat x^\theta_{t_0}+\sigma_{t_n}\epsilon$\end{tabular} & $(f_{\theta^{fake},n}-f_{\Phi,n})/w$ & $\hat x^\theta_{t_0}$ & $MSE(O_t,SG(O_t-O_s))$ \\
LongLive~\cite{yang2026longlive} & \begin{tabular}[c]{@{}l@{}}$\hat x^{\theta,k}_{t_0}=G_\theta(z_k;\mathcal H_k,c_k)$\\$\mathcal W_{k+1}=Keep_{W,sink}(\hat x^{\theta,\leq k}_{t_0})$\\$\mathcal H_{k+1}=$\\$\quad SG\!\left(\left\{\begin{array}{@{}l@{\;}l@{}}KV_\theta(\mathcal W_{k+1},c_{k+1}),&c_{k+1}\ne c_k\\Keep_{W,sink}(\mathcal H_k\oplus K_k),&c_{k+1}=c_k\end{array}\right.\right)$\\$K_k=KV_\theta(\hat x^{\theta,k}_{t_0},c_k)$\\$\hat x^{\theta,k}_{t_n}=\alpha_{t_n}\hat x^{\theta,k}_{t_0}+\sigma_{t_n}\epsilon$\end{tabular} & \begin{tabular}[c]{@{}l@{}}$(f_{\theta^{fake}}-f_\Phi)/w_k$\\$\text{evaluated at }(\hat x^{\theta,k}_{t_n},t_n)$\end{tabular} & $\hat x^{\theta,k}_{t_0}$ & $MSE(O_t,SG(O_t-O_s))$ \\
Salt~\cite{ge2026salt} & \begin{tabular}[c]{@{}l@{}}$\Psi_\theta^{t\to r}(x)=x-(t-r)v_\theta(x,t,c)$\\$b<r<t_n,\quad K\in\{2,4,8\}$\\$R_f^{low/ref}=\bar z_f^{low/ref}(\bar z_f^{low/ref})^T$\\$\|\bar z_{f,i}\|_2=1$\\$L_{align}=\frac1{FS^2}\sum_{fij}[d_{fij}-\delta]_+$\\$d_{fij}=|R_f^{low}(i,j)-R_f^{ref}(i,j)|$\end{tabular} & $\Psi_\theta^{t_n\to b}(x_{t_n})$ & \begin{tabular}[c]{@{}l@{}}$\Psi_\theta^{r\to b}(z_r)$\\$z_r=\Psi_\theta^{t_n\to r}(x_{t_n})$\end{tabular} & \begin{tabular}[c]{@{}l@{}}$L_{DMD}$\\$+\lambda^{SC}MSE(O_s,O_t)\mathbf1[K=8]$\\${}+\lambda^{align}L_{align}\mathbf1[K\in\{2,4\}]$\end{tabular} \\
\midrule
\rowcolor{black!9}
\multicolumn{5}{@{}l}{\hspace{3pt}\textbf{Adversarial Distillation}}\\
SDXL-Lightning~\cite{lin2024sdxl} & \begin{tabular}[c]{@{}l@{}}$\Phi^{(i+1)}\leftarrow\theta^{(i)}$\\$\theta^{(i+1)}\leftarrow\Phi^{(i+1)}$\end{tabular} & $DDIM(x_{t_n},t_n,t_{n-m};\theta)$ & $Solver(x_{t_n},t_n,t_{n-m};\Phi)$ & \begin{tabular}[c]{@{}l@{}}$Adv(O_s,O_t\mid x_{t_n},t_n,t_{n-m},c)$\\$\quad\longrightarrow Adv(O_s,O_t\mid t_{n-m},c)$\end{tabular} \\
ADD~\cite{sauer2025adversarial} & \begin{tabular}[c]{@{}l@{}}$x_s=\alpha_sx_0+\sigma_s\epsilon_s,\quad\hat x^\theta_{t_0}=f_\theta(x_s,s)$\\$\hat x^\theta_{t_n}=\alpha_{t_n}\hat x^\theta_{t_0}+\sigma_{t_n}\epsilon,\quad\epsilon\perp\epsilon_s$\end{tabular} & $\hat x^\theta_{t_0}$ & $f_\Phi(SG(\hat x^\theta_{t_n}),t_n)$ & \begin{tabular}[c]{@{}l@{}}$Adv(O_s,x_{t_0})$\\$+\lambda^{ADD}(t_n)MSE(O_s,O_t)$\end{tabular} \\
LADD~\cite{sauer2024fast} & \begin{tabular}[c]{@{}l@{}}$\hat x^\theta_{t_0}=G_\theta(z),\quad\hat x^\Phi_{t_0}\sim p_\Phi$\\$ x_t(y)=\alpha_ty+\sigma_t\epsilon$\end{tabular} & $\{B_{\Phi,j}(x_{t_n}(\hat x^\theta_{t_0}),t_n)\}_j$ & $\{B_{\Phi,j}(x_{t_n}(\hat x^\Phi_{t_0}),t_n)\}_j$ & $Adv(O_s,O_t)$ \\
NitroFusion~\cite{chen2024nitrofusion} & \begin{tabular}[c]{@{}l@{}}$\hat x^\theta_{t_0}=G_\theta(z),\quad\hat x^\Phi_{t_0}\sim p_\Phi$\\$\mathcal D_{t^*}\subset\mathcal D$\end{tabular} & $B_\Phi(x_{t^*}(\hat x^\theta_{t_0}),t^*)$ & $B_\Phi(x_{t^*}(\hat x^\Phi_{t_0}),t^*)$ & $-\sum_{H\in\mathcal D_{t^*}}H(O_s)$ \\
APT~\cite{lin2025diffusion} & \begin{tabular}[c]{@{}l@{}}$\hat x^\theta_{t_0}=z-v_\theta(z,t_N,c),\quad x_{t_0}\sim p_{data}$\\$D(x,c)=\mathbb E_t\widehat D(x,t,c)$\\$L_D^{aR1}=\|D(x_{t_0})-D(x_{t_0}+\sigma_R\epsilon)\|_2^2$\end{tabular} & $\hat x^\theta_{t_0}$ & $x_{t_0}$ & $Adv(O_s,O_t)$ \\
\bottomrule
\end{tabular}}
\par\vspace{2pt}
\begin{minipage}{\linewidth}\footnotesize
$f_{\Phi,n}$ and $f_{\theta^{fake},n}$ are clean-data predictions at $(\hat x^\theta_{t_n},t_n)$, with analogous batch index $j$. Teacher and auxiliary weights stay fixed. SiD, SiDA, and SIM retain input derivatives. In stopped MSE, $O_t=\hat x^\theta_{t_0}$ and $O_s$ is the correction. SwiftBrush and v2 weight the noise residual $\epsilon_{\Phi,n}-\epsilon_{\theta^{fake},n}=(\alpha_{t_n}/\sigma_{t_n})(f_{\theta^{fake},n}-f_{\Phi,n})$ by $w_{\rm VSD}$. CausVid shows the unweighted reverse-KL correction. Its Wan implementation uses the normalized difference, as do DMD2 and Self Forcing, with $w=\|O_t-f_\Phi\|_1/d$. LongLive and eDMD use $w_k=\|O_t-f_\Phi\|_1/d$ and $w_k=\|O_t-f_\Phi^{PT}\|_1/d$, respectively.
\par\vspace{2pt}
SiD uses $(\lambda_1^{SiD},\lambda_2^{SiD})=(1-\rho,1)w_t/\sigma_t^4$. SiDA uses $(\lambda^{SiDA}_{1,j},\lambda^{SiDA}_{2,j})=(1-\rho,1)\lambda_{sid}q_j$. Here $\rho$ weights bias correction and $Adv_B$ averages generator loss over fake samples and spatial discriminator outputs.
\par\vspace{2pt}
VA-TDM uses $a_{\tau|r}=(1-\tau)/(1-r)$, $b_{\tau|r}^2=\tau^2-a_{\tau|r}^2r^2$, $v_\tau^A=(y_\tau-\hat x_0)/\tau$, and $v_\tau^B=\dot a_{\tau|r}x_r+(\dot b_{\tau|r}/b_{\tau|r})(y_\tau-a_{\tau|r}x_r)$. Here $\bar v_{\theta,\tau}^F=\mathbb E[v_\tau^F\mid y_\tau]$ uses the fixed outer measure, and $(\pi_A,\pi_B)=(\lambda_m,1-\lambda_m)$. SwiftBrush v2 uses $C(x,c)=\langle E_{image}(Decode(x)),E_{text}(c)\rangle$.
\par\vspace{2pt}
$Keep_{W,sink}$ retains LongLive's local window and frame sink. Salt uses $2\to4$ and $4\to8$ reference rollouts and $L_{DMD}=MSE(O_t,SG(O_t-\Delta_{DM}))$, where $\Delta_{DM}$ is the stopped correction. APT's $L_D^{aR1}$ regularizes the discriminator.
\par\vspace{2pt}
SDXL-Lightning drops the initial-noise condition from its discriminator during fine-tuning and retains text conditioning in both stages. The one-step variant re-noises both $x_0$ predictions to the discriminator timestep.
\end{minipage}
\end{table*}

Diffusion distillation transfers a pretrained model to a student that uses fewer denoising steps. The teacher can supply individual transitions, trajectory endpoints, or scores of perturbed samples. Table~\ref{tab:distilloverview}, Parts~\hyperref[tab:distilloverview]{A} and~\hyperref[tab:distilloverview-b]{B}, compares the targets and losses across these families.

\subsubsection{Step Distillation}
Direct distillation regresses the endpoint of a full teacher trajectory in one student step~\cite{luhman2021knowledge}. Each target requires a full teacher solve. Progressive Distillation~\cite{salimans2022progressive} shortens that supervision interval by training one student step to replace two teacher steps, then repeating the procedure with the distilled model as the teacher. Guided Distillation~\cite{meng2023distillation} first conditions the student on the classifier-free guidance (CFG) scale~\cite{ho2022classifier} to absorb the two guidance evaluations into one pass, then applies progressive distillation. TRACT~\cite{berthelot2023tract} divides the trajectory into intervals and distills their transitions before reducing the number of intervals.

\subsubsection{Consistency Distillation}
\begin{figure*}[!t]
  \centering
  \includegraphics[width=\linewidth]{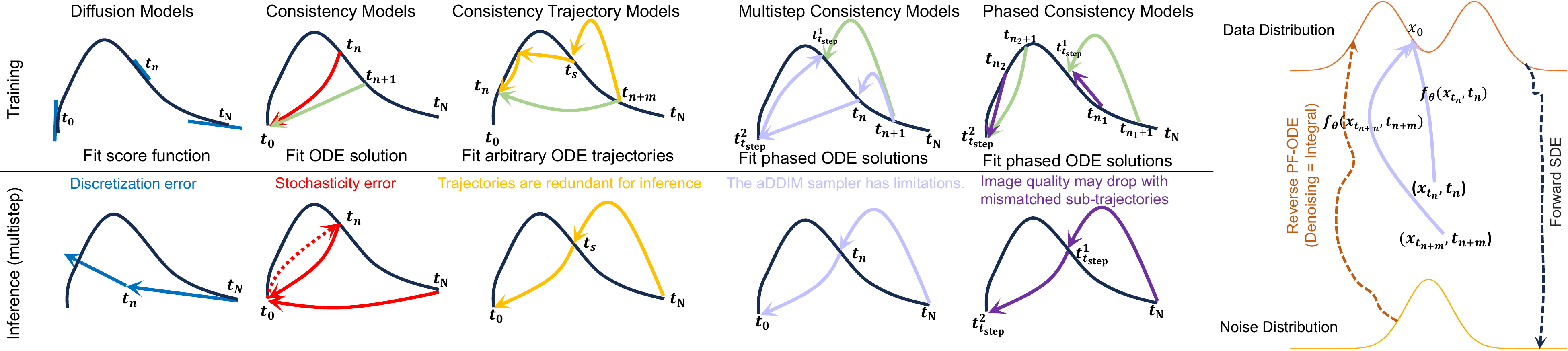}
  \caption{(Left) Consistency-distillation designs. (Right) Forward-SDE and reverse-time PF-ODE trajectories. Adapted from Phased Consistency Models (PCM)~\cite{wang2024phased}, Fig.~3.}
  \label{fig:consistency_distillation}
\end{figure*}

Consistency distillation (CD)~\cite{song2023consistency} trains the student to map points on the same PF-ODE trajectory to a common clean endpoint. Figure~\ref{fig:consistency_distillation} shows how later methods modify the target and divide the trajectory into shorter segments.

Convergence analyses link sampling error to the accuracy of the score and consistency estimates~\cite{lyu2023convergence}. The Latent Consistency Model (LCM)~\cite{luo2023latent} applies this objective to the compressed representation used by latent diffusion~\cite{rombach2022high}. LCM-LoRA~\cite{luo2023lcm} adapts the denoiser with low-rank updates, and PixArt-$\delta$~\cite{chen2024pixart} applies LCM to PixArt-$\alpha$. SCott~\cite{liu2024scott} uses an SDE solver to produce multistep targets and adds an adversarial loss on the resulting samples.

CTM~\cite{kim2023consistency} learns a two-time map \(G_\theta(x_t,t,s)\) that predicts an earlier state at time \(s\). PCM~\cite{wang2024phased} and MCM~\cite{heek2024multistep} train a shared network with consistency targets tied to segment boundaries. PCM adds adversarial supervision. Hyper-SD~\cite{ren2024hyper} combines segmented consistency distillation with human feedback. For text-to-video, T2V-Turbo~\cite{li2024t2v} trains the consistency student with image and video rewards. T2V-Turbo-v2~\cite{li2024t2v2} adds training-data selection and incorporates motion guidance into the teacher solver. TLCM~\cite{xie2024mlcm} uses data-free latent distillation and supplements inter-segment consistency with distribution, adversarial, and preference losses. SFD~\cite{zhou2024simple} studies fine-tuning at multiple times, variable evaluation counts, and the choice of teacher solver.

Initialization and checkpoint selection offer two ways to improve the consistency student. CMT~\cite{hu2026cmt} initializes consistency or mean-flow models by regressing clean endpoints from intermediate states on teacher trajectories. LCSC~\cite{liu2024linear} instead combines saved checkpoints as \(\sum_{i=1}^n a_i\theta_i\), with \(\sum_{i=1}^n a_i=1\).

VideoLCM~\cite{wang2023videolcm} applies CD to video, while AnimateLCM~\cite{wang2024animatelcm} distills appearance and motion priors separately. CCM~\cite{xiao2024ccm} transfers high-level control from a diffusion ControlNet and trains new control networks with consistency objectives or lightweight adapters.

\subsubsection{Adversarial Distillation}
\begin{figure}[!t]
  \centering
  \includegraphics[width=\linewidth]{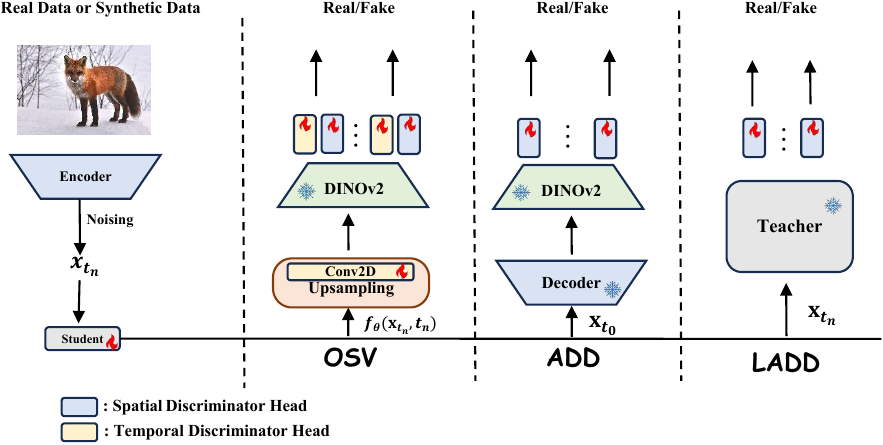}
  \caption{Discriminator designs for OSV, ADD, and LADD. Adapted from OSV~\cite{mao2024osv}, Fig.~3.}
  \label{fig:dis}
\end{figure}

Adversarial distillation uses a discriminator to compare student outputs with real data or teacher samples (Figure~\ref{fig:dis}). SDXL-Lightning~\cite{lin2024sdxl} conditions the discriminator on the initial noisy state during distillation and drops that condition during fine-tuning. It retains text conditioning in both stages. For one-step generation it predicts clean latents and re-noises both teacher and student outputs before discrimination. ADD~\cite{sauer2025adversarial} combines score distillation with pixel-space adversarial supervision, while LADD~\cite{sauer2024fast} uses features from a pretrained latent diffusion network. OSV~\cite{mao2024osv} uses real videos as the discriminator reference during pretraining and teacher-generated videos during distillation. NitroFusion~\cite{chen2024nitrofusion} selects discriminator heads from a pool according to the noise level.

AnimateDiff-Lightning~\cite{lin2024animatediff} applies progressive adversarial distillation to AnimateDiff~\cite{guo2023animatediff}. APT~\cite{lin2025diffusion} scales adversarial post-training to a one-step video generator. In blind super-resolution, AddSR~\cite{xie2024addsr} uses adversarial diffusion distillation to reduce the number of restoration steps.

\subsubsection{Flow Models}
\begin{figure*}[!t]
  \centering
  \includegraphics[width=\linewidth]{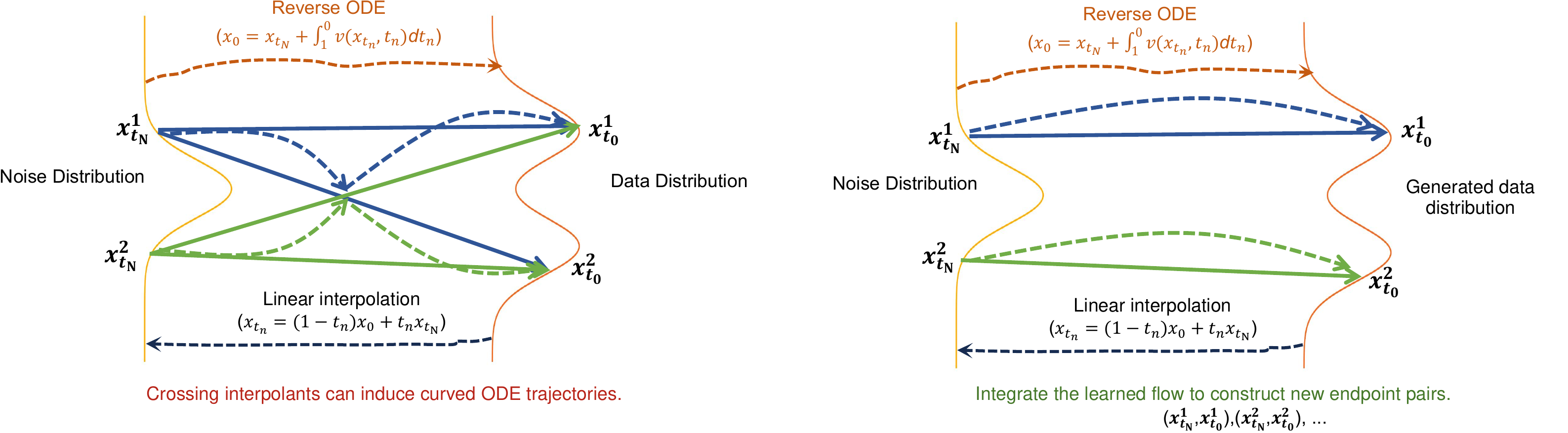}
\caption{Visualization of Rectified Flow~\cite{liu2022flow} Training and Inference. Reflow uses model-induced endpoint pairs to reduce trajectory curvature and improve coarse-step sampling.}
  \label{fig:flow}
\end{figure*}

Rectified Flow~\cite{liu2022flow} and Flow Matching~\cite{lipman2022flow} learn ODE velocities between base and data distributions. Linear interpolants in training can still produce curved learned trajectories. Reflow constructs a new endpoint coupling by integrating the current model and retraining on the resulting pairs (Figure~\ref{fig:flow}). InstaFlow~\cite{liu2023instaflow} applies this procedure to Stable Diffusion before distilling a one-step student. CFM~\cite{yang2024consistency} matches endpoint and velocity predictions along a flow. SlimFlow~\cite{zhu2025slimflow} uses annealed reflow and teacher velocity supervision to transfer the flow to a smaller network.

Flow-map distillation learns a finite-time transition \(G_\theta(z_t,t,s)\) from a pretrained vector field. Tong et al.~\cite{tong2026flowmap} sample only the base prior, match the teacher's generating velocity, and correct the student output distribution through its noising velocity. Their SiT-XL/2+REPA distillation reports one-step FID 1.45 on ImageNet \(256\times256\).

MeanFlow with a representation autoencoder~\cite{hu2026mfrae} trains the flow map in the autoencoder's latent space, using trajectory-aware initialization, teacher velocity supervision, and bootstrap refinement.

\subsubsection{Score Distillation}
\begin{figure}[!t]
  \centering
  \includegraphics[width=0.8\linewidth]{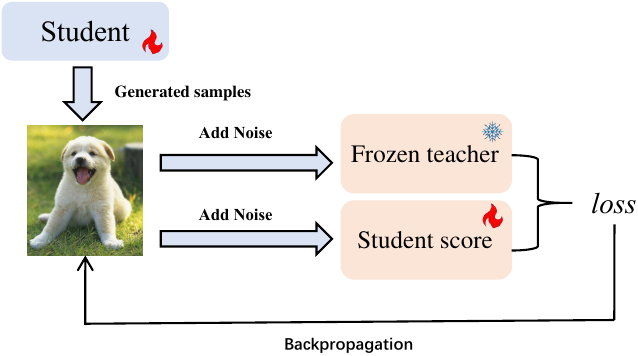}
\caption{Illustration of score distillation with a frozen teacher and an auxiliary score model for the student distribution.}
  \label{fig:score}
\end{figure}

Score Distillation Sampling originally used a pretrained text-to-image score to optimize 3D representations~\cite{poole2022dreamfusion}. In generator distillation, a teacher score and an auxiliary estimate of the student score guide the generator update (Figure~\ref{fig:score})~\cite{franceschi2024unifying}. 

SwiftBrush~\cite{nguyen2024swiftbrush} uses variational score distillation (VSD) to train a one-step generator. Its second version initializes from SD Turbo, trains with LoRA, and applies a clamped CLIP loss~\cite{dao2025swiftbrush}. In DMD~\cite{yin2024one}, the difference between the teacher and auxiliary scores estimates the gradient of a reverse KL on noisy samples. DMD2~\cite{yin2024improved} removes paired regression, updates the auxiliary score more often, and adds an adversarial loss with real images. SiD~\cite{zhou2024score} derives a score-based objective on perturbed student samples. SiDA adds a discriminator to this objective~\cite{zhou2024adversarial}, and TSD-SR applies score distillation to super-resolution~\cite{dong2024tsd}.

SIM~\cite{luo2025one} derives gradients of a score discrepancy for an implicit generator. For few-step sampling, eDMD~\cite{zhao2026edmd} fits an auxiliary score to perturbed final student outputs and reuses it at intermediate updates. VA-TDM~\cite{luo2026vatdm} applies transition-level supervision with a critic on velocity fields. VarFlow~\cite{shao2025varflow} estimates an energy distance from samples of the noisy student and reference distributions.

Video score distillation also changes which states the student sees during training. CausVid~\cite{yin2025causvid} distills a bidirectional teacher into a four-step causal student using independently noised chunks and temporal key--value caching. Self Forcing~\cite{huang2025selfforcing} rolls the student out on its own generated history. The DMD variant matches the resulting video distribution to the teacher. LongLive~\cite{yang2026longlive} extends rollout training to longer sequences with bounded context. Salt~\cite{ge2026salt} adds endpoint consistency to distribution matching and trains with the cache policy used at inference.

\subsubsection{Other Distillation Methods}
MSD~\cite{song2024multi} partitions a conditional model into multiple one-step students, each responsible for a subset of conditions.

Moment-matching distillation matches moments of stochastic multistep transitions, providing a distribution-level target for accelerated sampling~\cite{salimans2024moment}. Simplified Matching Distillation (SMD) regresses teacher velocities on student rollouts and uses a training-time residual proxy for within-interval variation~\cite{zhang2026onpolicy}.

\subsection{Consistency Training}
\begin{table*}[!b]
\centering
\setlength{\tabcolsep}{4pt}
\caption{Overview of Consistency Training.}
\label{tab:consistencyoverview}
\resizebox{\linewidth}{!}{
\begin{tabular}{@{}l l l@{}}
\toprule
\textbf{Model} & \textbf{Preliminaries} & \textbf{Loss Function} \\
\midrule
CT~\cite{song2023consistency} & $x_{t_n}=x_0+\sigma_{t_n}\epsilon,\quad x_{t_{n-m}}=x_0+\sigma_{t_{n-m}}\epsilon$ & $\lambda^{CM}(t_n)MSE(f_\theta(x_{t_n},t_n),f_{\theta^-}(x_{t_{n-m}},t_{n-m}))$ \\
iCT~\cite{song2023improved} & \begin{tabular}[c]{@{}l@{}}$x_{t_n}=x_0+\sigma_{t_n}\epsilon,\quad x_{t_{n-m}}=x_0+\sigma_{t_{n-m}}\epsilon$\\$\lambda^{\mathrm{iCT}}(t_n)=1/(\sigma_{t_n}-\sigma_{t_{n-m}})$\end{tabular} & $\lambda^{\mathrm{iCT}}(t_n)d_c(f_\theta(x_{t_n},t_n),f_{SG(\theta)}(x_{t_{n-m}},t_{n-m}))$ \\
TCM~\cite{lee2024truncated} & $x_{t_n}=x_0+\sigma_{t_n}\epsilon,\quad x_{t_{n-m}}=x_0+\sigma_{t_{n-m}}\epsilon$ & \begin{tabular}[c]{@{}l@{}}If $t_{n-m}<t^\prime$: $\lambda^{TCM}(t_n)d(f_\theta(x_{t_n},t_n),SG(f_{\theta_0^-}(x_{t_{n-m}},t_{n-m})))$,\\Else: $\lambda^{TCM}(t_n)d(f_\theta(x_{t_n},t_n),SG(f_\theta(x_{t_{n-m}},t_{n-m})))$\end{tabular} \\
ECT~\cite{geng2024consistency} & \begin{tabular}[c]{@{}l@{}}$\theta\leftarrow\Phi,\quad x_{t_n}=x_0+\sigma_{t_n}\epsilon,\quad x_{t_{n-m}}=x_0+\sigma_{t_{n-m}}\epsilon$\\$\Delta=f_\theta(x_{t_n},t_n)-SG(f_\theta(x_{t_{n-m}},t_{n-m})),$\\$\lambda^{ECT}=\bar w(t_n)/(\|SG(\Delta)\|_2^2+c^2)^p$\end{tabular} & $\lambda^{ECT}(t_n,\Delta)MSE(f_\theta(x_{t_n},t_n),f_{SG(\theta)}(x_{t_{n-m}},t_{n-m}))$ \\
\addlinespace[2pt]
sCM~\cite{lu2025scm} & \begin{tabular}[c]{@{}l@{}}$t=t_n,\quad x_t=\cos(t)x_0+\sin(t)z,\quad z\sim\mathcal N(0,\sigma_d^2I)$\\$\mathcal F^-_t=\mathcal F_{\theta^-}(x_t/\sigma_d,t),\quad g_t=\cos(t)\dot f_{\theta^-}$\\$\widetilde g_t=g_t/(\|g_t\|_2+c_{\rm tan})$\end{tabular} & \begin{tabular}[c]{@{}l@{}}$\frac{e^{w_\eta(t)}}{D}\bigl\|\mathcal F_\theta(x_t/\sigma_d,t)$\\$\qquad-SG(\mathcal F^-_t+\widetilde g_t)\bigr\|_2^2-w_\eta(t)$\end{tabular} \\
\addlinespace[2pt]
Shortcut~\cite{frans2024one} & \begin{tabular}[c]{@{}l@{}}$x_t=(1-t)\epsilon+tx_0,\quad t+2d\leq1,\quad x^\prime_{t+d}=x_t+d\,f_\theta(x_t,t,d)$\\$s_{target}=\tfrac12[f_\theta(x_t,t,d)+f_\theta(x^\prime_{t+d},t+d,d)]$\end{tabular} & \begin{tabular}[c]{@{}l@{}}$MSE(f_\theta(x_t,t,2d),SG(s_{target})),$\\$L_{d=0}=MSE(f_\theta(x_t,t,0),x_0-\epsilon)$\end{tabular} \\
MeanFlow~\cite{geng2025meanflow} & \begin{tabular}[c]{@{}l@{}}$z_{t_n}=(1-t_n)x_0+t_n\epsilon,\quad v=\epsilon-x_0,\quad r=t_{n-m}$\\$\frac{du}{dt_n}=(\partial_zu)v+\partial_{t_n}u\quad(r\ \mathrm{fixed})$\end{tabular} & $\lambda^{MF}(t_n,r)MSE(u_\theta(z_{t_n},r,t_n),SG(v-(t_n-r)\tfrac{d}{dt_n}u_\theta))$ \\
iMF~\cite{geng2026improvedmeanflow} & \begin{tabular}[c]{@{}l@{}}$z_{t_n}=(1-t_n)x_0+t_n\epsilon,\quad v=\epsilon-x_0,\quad r=t_{n-m}$\\$V_\theta=u_\theta+(t_n-r)SG(\mathcal D_{t_n}^{v_\theta}u_\theta),\quad v_\theta=u_\theta(z_{t_n},t_n,t_n)$\end{tabular} & $\lambda^{iMF}(t_n,r)MSE(V_\theta,v)$ \\
IMM~\cite{zhou2025imm} & $X,X^\prime\stackrel{iid}{\sim} p_{s|t_n}^\theta,\quad Y,Y^\prime\stackrel{iid}{\sim} p_{s|r}^{\theta^-},\quad s\leq r<t_n$ & $\lambda^{IMM}(s,t_n)\mathbb E[k(X,X^\prime)+k(Y,Y^\prime)-2k(X,Y)]$ \\
SCT~\cite{wang2024stable} & \begin{tabular}[c]{@{}l@{}}$\pi_i=q_{t_n}(x_{t_n}|x_0^i)/\sum_jq_{t_n}(x_{t_n}|x_0^j),$\\$\epsilon^{VR}=\sum_i\pi_i(x_{t_n}-\alpha_{t_n}x_0^i)/\sigma_{t_n}$\\$x_{t_{n-m}}=\frac{\alpha_{t_{n-m}}}{\alpha_{t_n}}(x_{t_n}-\sigma_{t_n}\epsilon^{VR})+\sigma_{t_{n-m}}\epsilon^{VR}$\end{tabular} & $\lambda^{SCT}(t_n)MSE(f_\theta(x_{t_n},t_n),f_{\theta^-}(x_{t_{n-m}},t_{n-m}))$ \\
\addlinespace[2pt]
FMM (direct)~\cite{boffi2025fmm} & $I_t=\alpha_tx_0+\beta_tx_1+\gamma_tz,\quad Z=f_\theta(I_t,t,s),\quad f_\theta(x,s,s)=x$ & $\lambda^{FMM}(s,t)[MSE(\partial_tf_\theta(Z,s,t),\dot I_t)+MSE(f_\theta(Z,s,t),I_t)]$ \\
AYT~\cite{kim2026ayt} & \begin{tabular}[c]{@{}l@{}}$x_{t_n}=x_0+\sigma_{t_n}\epsilon,\quad x_{t_{n-m}}=x_0+\sigma_{t_{n-m}}\epsilon$\\$\varphi=\arg\min_\varphi\mathbb E_{x,i,a}MSE(\varphi_i(T_a^i(x)),a)$\end{tabular} & $\lambda^{AYT}(t_n)MSE(\varphi(f_\theta(x_{t_n},t_n)),\varphi(SG(f_{\theta^-}(x_{t_{n-m}},t_{n-m}))))$ \\
TD diffusion~\cite{ying2026td} & \begin{tabular}[c]{@{}l@{}}$e_n=x_0-f_\theta(x_{t_n},t_n),\quad e^-_{n-1}=x_0-f_{\theta^-}(x_{t_{n-1}},t_{n-1})$\\$\delta_n=A_ne_n-A_{n-1}e^-_{n-1},\quad A_n=\frac{\alpha_{t_{n-1}}(1-\alpha_{t_n}^2/\alpha_{t_{n-1}}^2)}{1-\alpha_{t_n}^2},\quad n\geq2$\end{tabular} & $\|\delta_n\|_2^2+\lambda^{DDPM}(t_n)A_n^2MSE(f_\theta(x_{t_n},t_n),x_0)$ \\
\addlinespace[2pt]
Euler Mean Flows~\cite{li2026emf} & \begin{tabular}[c]{@{}l@{}}$z_{t_n}=(1-t_n)x_0+t_n\epsilon,\quad v=\epsilon-x_0,\quad r=t_{n-m},\quad\Delta=t_n-r,\quad h>0$\\$z^\prime=z_{t_n}-h\,SG(u_\theta(z_{t_n},t_n,t_n)),$\\$\delta_h^-u=\left\{\begin{array}{ll}\frac{u_\theta(z^\prime,r,t_n-h)-u_\theta(z_{t_n},r,t_n)}h,&\Delta>h\\0,&0\leq\Delta\leq h\end{array}\right.$\end{tabular} & \begin{tabular}[c]{@{}l@{}}$MSE(u_\theta(z_{t_n},r,t_n),v+(t_n-r-h)_+SG(\delta_h^-u))$\\$L_{r=t_n}=MSE(u_\theta(z_{t_n},t_n,t_n),v)$\end{tabular} \\
MF curriculum~\cite{kim2026meanflowtraining} & \begin{tabular}[c]{@{}l@{}}$z_{t_n}=(1-t_n)x_0+t_n\epsilon,\quad v=\epsilon-x_0,\quad r=t_{n-m}$\\$\rho_k=1-k/K,\quad\beta_k(\Delta)=1-\rho_k+\lambda\rho_k(1-\Delta),$\\$\Delta=t_n-r,\quad\lambda^{-1}=\mathbb E[1-\Delta]$\end{tabular} & \begin{tabular}[c]{@{}l@{}}$\mathbb E_{t_n\sim p_{acc}}\alpha(t_n)MSE(u_\theta(z_{t_n},t_n,t_n),v)$\\$\quad +\mathbb E_{r<t_n}\beta_k(t_n-r)L_{MF}$\end{tabular} \\
Discrete MF~\cite{hsu2026discretemf} & \begin{tabular}[c]{@{}l@{}}$z_{t_n}=(1-t_n)x_0+t_n\epsilon,\quad v=\epsilon-x_0,\quad r=t_{n-m},\quad0\leq r<t_n<1,\quad q>1$\\$b_0=v,\quad b_{K-1}=SG(v-(t_n-r)\mathcal D_{t_n}^vu_\theta)$\\$\chi(t)=t/(1-t),$\\$\delta_i=t_n-\chi^{-1}(\chi(t_n)-(\chi(t_n)-\chi(r))/q^i)$\end{tabular} & \begin{tabular}[c]{@{}l@{}}$b_i=\frac{v\delta_i+(t_n-r)SG(u_\theta(z_{t_n}-v\delta_i,r,t_n-\delta_i))}{\delta_i+t_n-r},\quad1\leq i\leq K-2$\\$MSE(u_\theta(z_{t_n},r,t_n),b_i)$\end{tabular} \\
Decoupled MF~\cite{lee2026decoupled} & \begin{tabular}[c]{@{}l@{}}$z_{t_n}=(1-t_n)x_0+t_n\epsilon,\quad v=\epsilon-x_0,\quad r=t_{n-m}$\\$u_\theta(z_{t_n},r,t_n)=Decoder_\theta(Encoder_\theta(z_{t_n},t_n),r)$\\$R=u_\theta-v+(t_n-r)SG(\mathcal D_{t_n}^vu_\theta)$\end{tabular} & \begin{tabular}[c]{@{}l@{}}$\mathbb E_{r<t_n}[\log(1+e^{-a_\eta(r,t_n)}MSE(R,0))+\tfrac12a_\eta(r,t_n)]$\\$\quad +L_{FM}^{Cauchy}$\\$L_{FM}^{Cauchy}=\mathbb E_{t_n}[\log(1+e^{-a_\eta(t_n,t_n)}MSE(u_\theta(z_{t_n},t_n,t_n),v))$\\$\quad +\tfrac12a_\eta(t_n,t_n)]$\end{tabular} \\
\bottomrule
\end{tabular}}
\par\smallskip
\begin{minipage}{\linewidth}\footnotesize
\textit{sCM.} In TrigFlow, $f_{\theta^-}(x_t,t)=\cos(t)x_t-\sigma_d\sin(t)\mathcal F^-_t$, with $\theta^-=SG(\theta)$. Along the training path,
$\dot f_{\theta^-}=\cos(t)(\dot x_t-\sigma_d \mathcal F^-_t)-\sin(t)(x_t+\sigma_d\dot{\mathcal F}_t^-)$ and
$\dot{\mathcal F}_t^-=\operatorname{JVP}_{(\dot x_t/\sigma_d,1)}\mathcal F_{\theta^-}(x_t/\sigma_d,t)$.
For consistency training, $\dot x_t=\cos(t)z-\sin(t)x_0$. The loss uses sample dimension $D$, learned log-weight $w_\eta(t)$ and tangent-normalization constant $c_{\rm tan}>0$. The optional warmup multiplies the second term of $\dot f_{\theta^-}$ by a coefficient increasing from zero to one.
\end{minipage}
\end{table*}

Distillation uses a pretrained teacher to supply training targets. Computing those targets adds training cost and can require both models in GPU memory. Consistency Training (CT)~\cite{song2023consistency} learns from pairs of noisy data points that share the same perturbation. In the variance-exploding parameterization, the score can be expressed using the data and noisy state:
\begin{equation}
\nabla \log p_{t_n}({x}_{t_n}) = -\mathbb{E} \left[ \frac{{x}_{t_n} - {x}}{\sigma_{t_n}^2} \middle| {x}_{t_n} \right]
\end{equation}
The conditional perturbation score is \(\nabla_{x_{t_n}}\log q(x_{t_n}\mid x)=-(x_{t_n}-x)/\sigma_{t_n}^2\). Its expectation conditional on \(x_{t_n}\) gives the marginal score above. CT uses the same \(x\) and noise \(\epsilon\) at adjacent noise levels:
\begin{equation}
\lambda(t_{n}) MSE \left( {f}_{\theta}({x} + \sigma_{t_{n+1}}{\epsilon}, t_{n+1}), \operatorname{sg}({f}_{\theta^-}({x} + \sigma_{t_n} {\epsilon}, t_n)) \right) 
\end{equation}

Improved Consistency Training (iCT)~\cite{song2023improved} uses the current network weights for the stopped target and a pseudo-Huber discrepancy:
\begin{equation}
\lambda(t_{n}) d_c \left( {f}_{\theta}({x} + \sigma_{t_{n+1}}{\epsilon}, t_{n+1}), \operatorname{sg}({f}_{\theta}({x} + \sigma_{t_n} {\epsilon}, t_n)) \right) 
\end{equation}

iCT adjusts the noise schedule and grows the number of discretization intervals during training. It weights adjacent pairs by inverse noise-level spacing, \(\lambda(\sigma_{t_n})=1/(\sigma_{t_{n+1}}-\sigma_{t_n})\). TCM~\cite{lee2024truncated} restricts training to a truncated time range for one-step generation, whereas ECT~\cite{geng2024consistency} starts from a pretrained diffusion model and uses adaptive weights. SCT~\cite{wang2024stable} uses stage-wise tuning to reduce target variance and discretization error. sCM~\cite{lu2025scm} takes a continuous-time approach, constructing a normalized tangent target along a trigonometric path.

Shortcut models~\cite{frans2024one} condition on the requested step size \(d\) and match one large update to two successive smaller ones:
\begin{equation}
    \label{eq:shortcut_loss}
\left\|f_\theta(x_t,t,2d)-\operatorname{sg}(s_{\rm target})\right\|_2^2,
\end{equation}
For \(d>0\) and \(t+2d\leq1\), \(s_{\rm target}=\tfrac12[f_\theta(x_t,t,d)+f_\theta(x_{t+d}',t+d,d)]\), where \(x_{t+d}'=x_t+d f_\theta(x_t,t,d)\). Training samples valid sizes from \(d\in\{1/128,1/64,\dots,1/2,1\}\); the \(d=0\) branch uses flow matching.

MeanFlow~\cite{geng2025meanflow} learns the interval-average velocity directly, so multiplying it by the interval length gives a finite update. Improved MeanFlow~\cite{geng2026improvedmeanflow} supervises instantaneous velocity during training and conditions the network on guidance. The cost and conditioning of the average-velocity target motivate several variants. Euler Mean Flows~\cite{li2026emf} approximate the total derivative by a finite difference, while the curricula of Kim et al.~\cite{kim2026meanflowtraining} and Hsu and Wood~\cite{hsu2026discretemf} organize training around easier velocity or discrete consistency targets. Decoupled MeanFlow~\cite{lee2026decoupled} confines destination-time conditioning to the final transformer blocks.

Other objectives change the quantity being matched. Inductive Moment Matching~\cite{zhou2025imm} matches distributional moments, while Flow Map Matching~\cite{boffi2025fmm} derives objectives for two-time maps from stochastic interpolants. AYT~\cite{kim2026ayt} aligns consistency tangents in feature space, and TD diffusion~\cite{ying2026td} relates discrepancies at adjacent timesteps. Table~\ref{tab:consistencyoverview} compares the targets and losses.

\begin{figure*}[!t]
  \centering
  \includegraphics[width=0.9\linewidth]{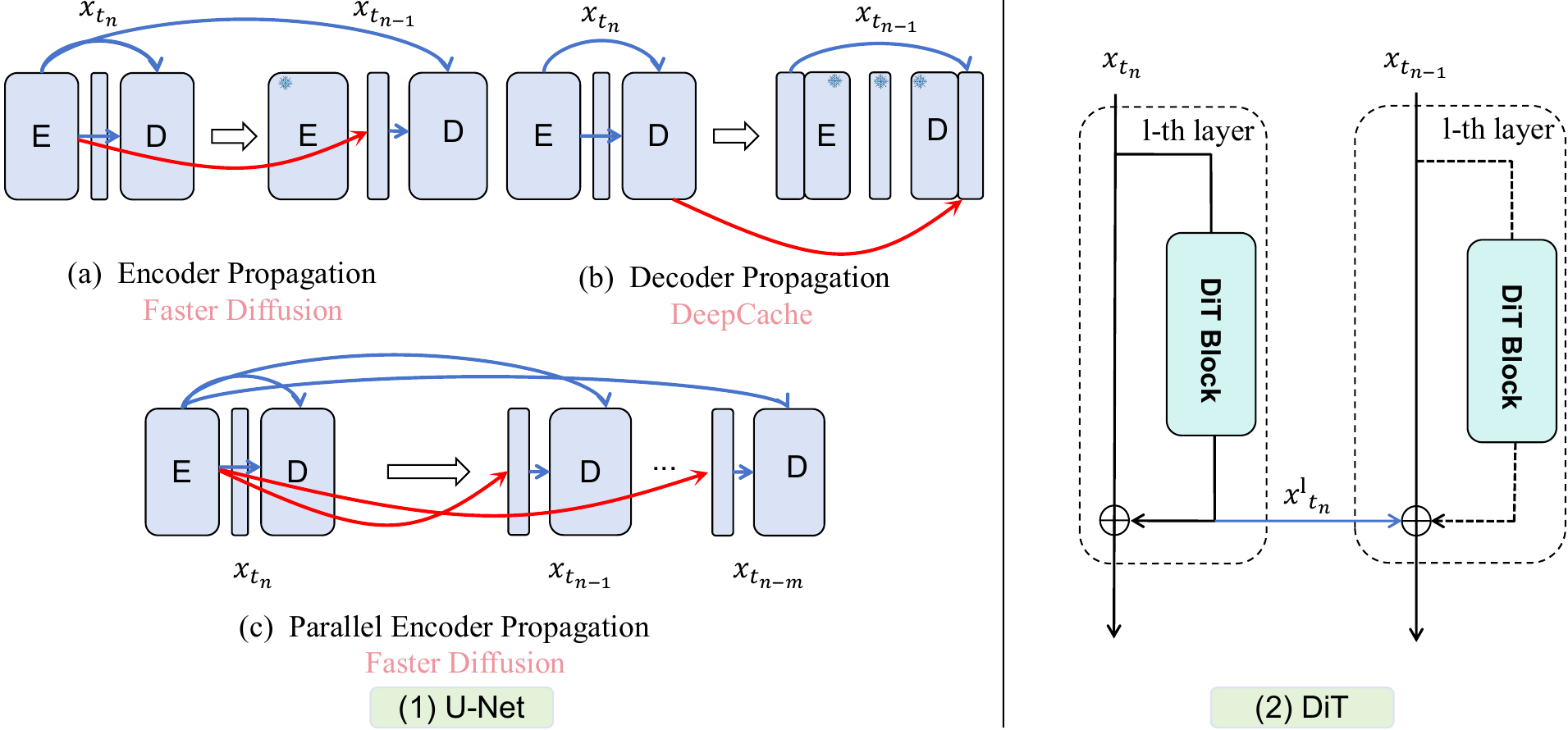}
  \caption{Cache-based Acceleration.}
  \label{fig:cache}
\end{figure*}

\subsection{Sampling Methods}
\begin{table*}[!b]
\centering
\setlength{\tabcolsep}{4pt}
\caption{Overview of Sampling Methods.}
\label{tab:sample}
\resizebox{\linewidth}{!}{
\begin{tabular}{@{}l l l@{}}
\toprule
\textbf{Model} & \textbf{Preliminaries} & \textbf{Reverse Process} \\
\midrule
DDIM~\cite{song2020denoising} & $x_{t_n}$ & $x_{t_{n-m}}=\alpha_{t_{n-m}}\frac{x_{t_n}-\sigma_{t_n}\epsilon_\theta(x_{t_n},t_n)}{\alpha_{t_n}}+\sigma_{t_{n-m}}\epsilon_\theta(x_{t_n},t_n)$ \\
Heun~\cite{karras2022elucidating} & $\alpha_t=1,\quad x_{t_n},\quad\hat x_{t_{n-m}}=x_{t_n}-\epsilon_\theta(x_{t_n},t_n)(\sigma_{t_n}-\sigma_{t_{n-m}})$ & \begin{tabular}[c]{@{}l@{}}$x_{t_{n-m}}=x_{t_n}-\tfrac12[\epsilon_\theta(\hat x_{t_{n-m}},t_{n-m})+\epsilon_\theta(x_{t_n},t_n)](\sigma_{t_n}-\sigma_{t_{n-m}}),$\\$\quad\sigma_{t_{n-m}}>0$\\$x_{t_{n-m}}=\hat x_{t_{n-m}},\quad\sigma_{t_{n-m}}=0$\end{tabular} \\
\addlinespace[2pt]
DPM-Solver-2~\cite{lu2022dpm} & \begin{tabular}[c]{@{}l@{}}$\lambda_{t_n}=\log(\alpha_{t_n}/\sigma_{t_n}),\quad h=\lambda_{t_{n-m}}-\lambda_{t_n},$\\$s_1=t_\lambda(\lambda_{t_n}+r_1h),\quad r_1=0.5$\end{tabular} & \begin{tabular}[c]{@{}l@{}}$\hat u=\frac{\alpha_{s_1}}{\alpha_{t_n}}x_{t_n}-\sigma_{s_1}(e^{r_1h}-1)\epsilon_\theta(x_{t_n},t_n)$\\$x_{t_{n-m}}=\frac{\alpha_{t_{n-m}}}{\alpha_{t_n}}x_{t_n}-\sigma_{t_{n-m}}(e^h-1)\epsilon_\theta(\hat u,s_1)$\end{tabular} \\
DPM-Solver++(2S)~\cite{lu2022dpm1} & \begin{tabular}[c]{@{}l@{}}$\lambda_{t_n}=\log(\alpha_{t_n}/\sigma_{t_n}),\quad h=\lambda_{t_{n-m}}-\lambda_{t_n},$\\$s_1=t_\lambda(\lambda_{t_n}+r_1h),\quad r_1=0.5$\\$f_\theta(x,t)=(x-\sigma_t\epsilon_\theta(x,t))/\alpha_t$\end{tabular} & \begin{tabular}[c]{@{}l@{}}$\hat u=\frac{\sigma_{s_1}}{\sigma_{t_n}}x_{t_n}-\alpha_{s_1}(e^{-r_1h}-1)f_\theta(x_{t_n},t_n),$\\$D=(1-\frac1{2r_1})f_\theta(x_{t_n},t_n)+\frac1{2r_1}f_\theta(\hat u,s_1)$\\$x_{t_{n-m}}=\frac{\sigma_{t_{n-m}}}{\sigma_{t_n}}x_{t_n}-\alpha_{t_{n-m}}(e^{-h}-1)D$\end{tabular} \\
UniP-2~\cite{zhao2024unipc} & \begin{tabular}[c]{@{}l@{}}$\lambda_{t_n}=\log(\alpha_{t_n}/\sigma_{t_n}),\quad h=\lambda_{t_{n-m}}-\lambda_{t_n}$\\$s_1=t_{n+m}\ \text{(history)},\quad r_1=(\lambda_{t_{n+m}}-\lambda_{t_n})/h<0$\\$a_1=(e^h-1-h)/(hB(h))$\end{tabular} & \begin{tabular}[c]{@{}l@{}}$D_1=\epsilon_\theta(x_{s_1},s_1)-\epsilon_\theta(x_{t_n},t_n)$\\$x_{t_{n-m}}=\frac{\alpha_{t_{n-m}}}{\alpha_{t_n}}x_{t_n}-\sigma_{t_{n-m}}(e^h-1)\epsilon_\theta(x_{t_n},t_n)$\\$\quad-\sigma_{t_{n-m}}B(h)a_1D_1/r_1$\end{tabular} \\
AMED~\cite{zhou2024fast} & \begin{tabular}[c]{@{}l@{}}$\alpha_t=1,\quad x_{t_n},\quad(s_n,c_n)=g_\psi(B_\theta(x_{t_n},t_n),t_n,t_{n-m}),$\\$t_{n-m}<s_n<t_n$\end{tabular} & \begin{tabular}[c]{@{}l@{}}$x_{s_n}=x_{t_n}+(s_n-t_n)\epsilon_\theta(x_{t_n},t_n)$\\$x_{t_{n-m}}=x_{t_n}+c_n(t_{n-m}-t_n)\epsilon_\theta(x_{s_n},s_n)$\end{tabular} \\
\addlinespace[2pt]
AYS~\cite{sabour2024ays} & $R_t=f_{cont}(X_{0:t},t)-f_{disc}(X_{0:t},t;\mathcal T)$ & \begin{tabular}[c]{@{}l@{}}$\mathcal T^*=\arg\min_{\mathcal T}\sum_i\tfrac12\mathbb E\int_{t_{i-1}}^{t_i}\|R_t\|_2^2/g(t)^2\,dt$\\$x_{t_{i-1}}=Solver(x_{t_i},t_i,t_{i-1};\theta),\quad t_i\in\mathcal T^*$\end{tabular} \\
PCS / PCS+ft~\cite{wu2026timestep} & \begin{tabular}[c]{@{}l@{}}$m_j\sim Bern(s_j),\quad0\leq s_j\leq1,\quad\|s\|_1\leq K$\\$\theta^*(m)=\theta_0\ (PCS)$\end{tabular} & \begin{tabular}[c]{@{}l@{}}$s^*=\arg\min_s\mathbb E_m d(G_{\theta^*(m)}(z;m),G_{\theta_0}(z;\mathbf1))$\\$\theta^*(m)=\arg\min_\theta\mathbb E_{t\sim m}\lambda(t)MSE(\epsilon_\theta(x_t,t),\epsilon)\ (PCS+ft)$\end{tabular} \\
TADA~\cite{chen2025tada} & \begin{tabular}[c]{@{}l@{}}$X_t=(x_t^{(0)},\ldots,x_t^{(N-1)}),\quad y_t=r_t^TX_t,\quad r_t=\Sigma_t^{-1}\mu_t/(\mu_t^T\Sigma_t^{-1}\mu_t)$\\$\sigma_{eff,t}^2=(\mu_t^T\Sigma_t^{-1}\mu_t)^{-1},\quad f_\theta(y,\sigma)=D_\theta^{VE}(y,\sigma)$\end{tabular} & \begin{tabular}[c]{@{}l@{}}$\dot x_t^{(j)}=x_t^{(j+1)}\ (0\leq j<N-1)$\\$\dot x_t^{(N-1)}=\frac{N!}{(1-t)^N}\left[f_\theta(y_t,\sigma_{eff,t})-\sum_{j=0}^{N-1}\frac{(1-t)^j}{j!}x_t^{(j)}\right]$\end{tabular} \\
\addlinespace[2pt]
Secant integration~\cite{liu2025secant} & \begin{tabular}[c]{@{}l@{}}$u\sim U(t_{n-m},t_n)$\\$\hat x_u=x_{t_n}+(u-t_n)S_{SG(\theta)}(x_{t_n},t_n,u)$\\$L=MSE(S_\theta(x_{t_n},t_n,t_{n-m}),\Phi(\hat x_u,u;\phi))$\end{tabular} & $x_{t_{n-m}}=x_{t_n}+(t_{n-m}-t_n)S_\theta(x_{t_n},t_n,t_{n-m})$ \\
LTC-Accel~\cite{zhu2025ltc} & $\Delta_n=x_{t_{n-m}}-x_{t_n},\quad\gamma_n=\frac{\varrho(t_{n-m})-\varrho(t_n)}{\varrho(t_n)-\varrho(t_{n+m})}$ & \begin{tabular}[c]{@{}l@{}}$\hat w_n=\left.\frac{\Delta_n^T\Delta_{n+m}}{\gamma_n\|\Delta_{n+m}\|_2^2}\right|_{cal}$\\$\hat x_{t_{n-m}}=x_{t_n}+\hat w_n\gamma_n\Delta_{n+m}$\end{tabular} \\
F-DPMSolver~\cite{jiao2026fdpm} & $x_{t_n},\quad\rho_n=\sigma_{t_{n-m}}/\sigma_{t_n}$ & \begin{tabular}[c]{@{}l@{}}$\hat x_{t_{n-m}}=DDIM(x_{t_n},t_n,t_{n-m};\theta)$\\$x_{t_{n-m}}=\rho_nx_{t_n}+(\alpha_{t_{n-m}}-\rho_n\alpha_{t_n})f_\theta(\hat x_{t_{n-m}},t_{n-m})$\end{tabular} \\
GeoRK2~\cite{sun2026geork2} & \begin{tabular}[c]{@{}l@{}}$P_n=U_nU_n^T,\quad h_m=P_n(h_n+\tfrac{\Delta t}{2}v_n),$\\$v_m=(\mathcal B(h_m)-h_m)/\Delta t$\end{tabular} & \begin{tabular}[c]{@{}l@{}}$h_p=P_n(h_n+\Delta t\,v_m)$\\$h_{n+1}=h_p-\lambda\bar G_n^{-1}(h_p-\mathcal B(h_p))$\end{tabular} \\
\addlinespace[2pt]
Spectral schedules~\cite{esteves2026spectral} & \begin{tabular}[c]{@{}l@{}}$\kappa_t=\kappa_{max}^t\kappa_{min}^{1-t},\quad\mu_F=N_f+(1-N_f)t,\quad\mu_P=F_{x_0}^{-1}(1-t)$\\$\lambda_{F/P}=-\log\kappa_t-\log\widetilde\Psi_{x_0}(\mu_{F/P}),$\\$F_{x_0}(q)=\frac{\int_1^q\widetilde\Psi_{x_0}(u)du}{\int_1^{N_f}\widetilde\Psi_{x_0}(u)du}$\end{tabular} & \begin{tabular}[c]{@{}l@{}}$\lambda_t=\tfrac14(\lambda_F+\lambda_P),$\\$\alpha_t^2=sigmoid(2\lambda_t),\quad\sigma_t^2=sigmoid(-2\lambda_t)$\end{tabular} \\
Q-Sched~\cite{frumkin2026qsched} & \begin{tabular}[c]{@{}l@{}}$\epsilon_n^Q=\epsilon_\theta^Q(x_{t_n},t_n),\quad 0\leq s'\leq t_{n-m}<t_n$\\$c^*=\arg\min_c\mathbb E[L_{TC}(G^c(z))+kL_{IQ}(G^c(z))]$\end{tabular} & \begin{tabular}[c]{@{}l@{}}$\widetilde x_0=(c_n^x x_{t_n}-\sigma_{t_n}c_n^\epsilon\epsilon_n^Q)/\alpha_{t_n}$\\$\eta_n=\sqrt{1-\alpha_{t_{n-m}}^2/\alpha_{s'}^2},\quad z\sim\mathcal N(0,I)$\\$x_{t_{n-m}}=\frac{\alpha_{t_{n-m}}}{\alpha_{s'}}\bigl(\alpha_{s'}\widetilde x_0+\sigma_{s'}c_n^\epsilon\epsilon_n^Q\bigr)+\eta_n z$\end{tabular} \\
Collocation (theory)~\cite{gatmiry2026collocation} & \begin{tabular}[c]{@{}l@{}}$H=t_n-t_{n-m},\quad\Phi_{rev}(x,s)=-\Phi(x,t_n-s;\theta),\quad0\leq c_j\leq H$\\$\ell_j(c_i)=\mathbf1[i=j],\quad A_{ij}=\int_0^{c_j}\ell_i(s)ds,\quad X^{(0)}=x_{t_n}\mathbf1^T$\\$\mathcal F_c(X)_{:,j}=\Phi_{rev}(X_{:,j},c_j)$\end{tabular} & \begin{tabular}[c]{@{}l@{}}$X^{(k+1)}=x_{t_n}\mathbf1^T+\mathcal F_c(X^{(k)})A$\\$\hat x_{t_{n-m}}=x_{t_n}+\sum_j\Phi_{rev}(X_{:,j}^{(K)},c_j)\int_0^H\ell_j(s)ds$\end{tabular} \\
\bottomrule
\end{tabular}}
\end{table*}

DDIM~\cite{song2020denoising} discretizes a deterministic reverse trajectory. With a small evaluation budget, both the time grid and how the solver approximates the trajectory between grid points affect sampling quality.

\subsubsection{Accelerated Sampling}
In Table~\ref{tab:sample}, \(t_{n-m}<t_n\), \(t_\lambda\) is the inverse of \(\lambda_t\), and \(f_\theta\) denotes clean-data prediction in the sampling formulas. Half-log-SNR updates use positive signal and noise scales; UniP-2 uses the cached evaluation at $s_1=t_{n+m}>t_n$ from the preceding reverse step. It starts with a first-order update when this history is unavailable.

Heun corrects an Euler predictor with a second evaluation (Table~\ref{tab:sample}). DPM-Solver~\cite{lu2022dpm} integrates the linear part of the diffusion ODE analytically and expands the noise predictor along the trajectory in half-log-SNR time, $\lambda_t=\log(\alpha_t/\sigma_t)$:
\begin{equation}
\label{eq:dpm-taylor}
\begin{aligned}
x_{t_{n-m}} &= \frac{\alpha_{t_{n-m}}}{\alpha_{t_n}}x_{t_n}\\[-1pt]
&\quad-\alpha_{t_{n-m}}\sum_{i=0}^{k-1}
  \varepsilon_\theta^{(i)}(x_{\lambda_{t_n}},\lambda_{t_n})\\[-1pt]
&\quad\times\int_{\lambda_{t_n}}^{\lambda_{t_{n-m}}}
  e^{-\lambda}\frac{(\lambda-\lambda_{t_n})^i}{i!}\,\mathrm d\lambda
  +\mathcal{O}(|h|^{k+1}).
\end{aligned}
\end{equation}
where $h=\lambda_{t_{n-m}}-\lambda_{t_n}$, \(k\) is the solver order and \(\varepsilon_\theta^{(i)}\) is the \(i\)-th total derivative along the solution in half-log-SNR time. The integral part can be approximated at multiple orders as illustrated in Table~\textcolor{red}{~\ref{tab:sample}}. DDIM essentially corresponds to the first-order form of DPM-Solver. DPM-Solver++~\cite{lu2022dpm1} offers advantages over DPM-Solver in guided sampling. UniPC~\cite{zhao2024unipc} develops a unified corrector (UniC) that enhances accuracy without extra model evaluations after any existing DPM sampler, and derives a unified predictor (UniP) supporting arbitrary order as a byproduct. 

AMED~\cite{zhou2024fast} learns an approximation of the mean update direction over each integration interval, motivated by observed low-dimensional trajectory structure. Specifically,
\begin{equation}
\resizebox{\linewidth}{!}{$
x_r=x_t+\int_t^r F_\theta(x_a,a)\,\mathrm da,\qquad \overline{F}_{t,r}=\frac{1}{r-t}\int_t^r F_\theta(x_a,a)\,\mathrm da.
$}
\end{equation}
AMED trains a lightweight predictor with the denoiser fixed to approximate this mean direction using an intermediate evaluation.

OTS~\cite{xue2024accelerating} chooses timesteps for a fixed solver by minimizing discretization error with a constrained trust-region method. GITS~\cite{chen2024trajectory} uses kernel density estimation~\cite{fukunaga1975estimation} to explain the linear--nonlinear--linear shape of denoising trajectories and dynamic programming to place the timesteps.
Align Your Steps optimizes the time grid jointly with the model and sampler~\cite{sabour2024ays}. Wu et al.~\cite{wu2026timestep} connect inference grids with the timesteps used during training. Spectral schedules use image spectra to choose noise ranges~\cite{esteves2026spectral}, while Q-Sched calibrates update coefficients for a fixed quantized denoiser~\cite{frumkin2026qsched}.
Several solvers change how each interval is integrated. Secant integration learns an interval-average derivative from interior teacher evaluations~\cite{liu2025secant}, while LTC-Accel estimates an increment using neighboring transitions~\cite{zhu2025ltc}. F-DPMSolver supplies a lookahead correction by approximating a forward evaluation~\cite{jiao2026fdpm}. TADA enlarges the dynamical state with momentum variables~\cite{chen2025tada}, and GeoRK2 incorporates feature covariance into a second-order update~\cite{sun2026geork2}. Collocation gives integration bounds when the solver only has access to an approximate score~\cite{gatmiry2026collocation}.

\subsubsection{Parallel Sampling}

Parallel sampling shortens the sequential trajectory. DSNO~\cite{zheng2023fast} learns a neural operator that predicts the diffusion ODE trajectory at multiple times in one network pass. ParaDiGMS~\cite{shih2024parallel} keeps the underlying solver and iterates on batches of tentative denoising states. SRDS~\cite{selvam2024self} initializes the trajectory with a coarse solve, then refines its intervals in parallel and corrects their endpoints.
StreamDiffusionV2 coordinates block scheduling, rolling key--value caches and multi-GPU pipelining for live video~\cite{feng2026streamdiffusionv2}. Its pipeline is organized around first-frame latency and frame deadlines as well as steady-state throughput. Temporal key--value caching preserves attention context across video chunks, while feature caching across noise levels reuses denoiser computation within the sampling trajectory.
\subsection{Denoising Network Design}
Denoiser optimization reduces the work in a single forward evaluation. The main approaches modify attention, reuse activations, quantize computation, or prune the network.
\subsubsection{Efficient Attention Design}
Latent diffusion~\cite{rombach2022high} reduces the number of spatial positions on which the denoiser operates. U-DiT~\cite{tian2024u} uses a U-shaped transformer with token downsampling to reduce attention cost at high resolutions.

Standard self-attention scales quadratically with the number of latent tokens. SANA~\cite{xie2024sana} combines linear attention~\cite{katharopoulos2020transformers} with a $32\times$ spatially compressive autoencoder to reduce both token count and attention cost. Diffusion-RWKV~\cite{fei2024diffusion} adapts recurrent RWKV blocks~\cite{peng2023rwkv} to the denoiser. DiM~\cite{teng2024dim} adapts Mamba state-space layers~\cite{gu2023mamba}, scanning patch sequences in multiple directions to capture two-dimensional context.
Sparse Video-Gen identifies heads with spatial or temporal attention patterns~\cite{xi2025svg}. Sparse VideoGen2 instead groups semantically similar tokens and reorders them so that sparse operations form contiguous blocks for GPU execution~\cite{yang2025svg2}. Radial Attention builds its pattern from spatiotemporal decay~\cite{li2025radial}. RAPID carries attention-block importance across denoising steps so that it can reuse sparsity information~\cite{lin2026rapid}.

\ImageComparisonTable
\VideoComparisonTable

\subsubsection{Cache-based Acceleration}

DeepCache~\cite{ma2024deepcache} caches intermediate U-Net features from a full denoiser call. At an adjacent step it updates the shallow path and reuses the cached deep features (Figure~\ref{fig:cache}). Faster Diffusion~\cite{li2023faster} exploits the slower variation of encoder features across steps. Reusing them skips encoder evaluations and lets the decoder process several steps in parallel.

In a DiT, storing the output \(x_{t_n}^l\) of block \(l\) lets a later step reuse it without running the block again. FORA~\cite{selvaraju2024fora} refreshes attention and MLP outputs at fixed intervals, whereas L2C~\cite{ma2024learning} learns which layers to skip at each step. ToCa~\cite{zou2024accelerating} uses token sensitivity to select what to reuse within a layer, and BWCache~\cite{cui2026bwcache} makes separate reuse decisions for each transformer block.

The refresh signal determines when reuse becomes too costly in quality. AdaCache~\cite{kahatapitiya2024adaptive} adjusts refresh intervals using video motion. TeaCache~\cite{liu2025teacache} and MagCache~\cite{ma2025magcache} measure changes in timestep-modulated inputs and relative residual magnitude, respectively, while SenCache~\cite{haghighi2026sencache} estimates sensitivity to changes in both the latent and timestep. TaylorSeer~\cite{liu2025taylorseer} changes what is reused by extrapolating activations with finite-difference derivatives. ARCache~\cite{nan2026arcache} uses generation history to guide refresh and residual correction in autoregressive video.

\subsubsection{Quantization and Pruning}
1.58-bit FLUX~\cite{yang20241} constrains FLUX.1-dev weights to the ternary set \(\{-1,0,+1\}\), with nominal information content \(\log_2 3\approx1.58\) bits per weight. Fine detail can degrade at high resolutions. For video, VDMini~\cite{wu2024individual} prunes shallow blocks while retaining deep layers for motion and uses adversarial training. LD-Pruner~\cite{castells2024ld} uses shifts in latent statistics to score operators for structured pruning.
Q-Diffusion calibrates post-training quantization across diffusion timesteps~\cite{li2023qdiffusion}. SVDQuant absorbs outliers into a high-precision low-rank branch and fuses its execution with a four-bit branch in Nunchaku~\cite{li2025svdquant}. Attention kernels can also quantize the matrix products within attention. SageAttention2 quantizes the query--key product to INT4 and uses FP8 for the attention-weight--value product, together with smoothing and accumulation corrections~\cite{zhang2025sage2}.

\section{Quantitative Comparison}
\label{sec:quantitative}
Table~\ref{tab:quantitative} brings together published benchmarks on few-step image generation, sampler schedules, feature caching, and streaming video. Each panel identifies the source and evaluation protocol.

\subsection{Metrics and Evaluation Protocols}
\textbf{Image generation.} Fr\'echet Inception Distance (FID) compares the Gaussian statistics of real and generated images in an Inception feature space~\cite{heusel2017fid}. Spatial FID (sFID) uses intermediate spatial features to capture finer spatial statistics~\cite{nash2021sparse}. Lower values indicate closer feature distributions. Inception Score (IS) combines confident class predictions with diversity in the marginal class distribution; higher values are preferred~\cite{salimans2016improved}. The patch FID in panel (b) applies FID to $299\times299$ center crops following DMD2~\cite{yin2024improved}. CLIP-based alignment measures image--text similarity in a shared feature space~\cite{radford2021clip}; panel (b) follows DMD2's OpenCLIP-G evaluation~\cite{cherti2023openclip,yin2024improved}.

\textbf{Video fidelity and generation.} LPIPS measures perceptual distance in learned feature representations~\cite{zhang2018perceptual}. PSNR expresses pixel reconstruction error on a logarithmic scale, while SSIM combines luminance, contrast, and structural comparisons~\cite{wang2004ssim}. Lower LPIPS and higher PSNR/SSIM indicate greater fidelity. VBench separates generation quality and semantic alignment across its evaluation dimensions~\cite{huang2024vbench}. VBench-Long, part of VBench++, extends evaluation to long videos~\cite{huang2026vbenchpp}; panel (f) follows LongLive's official-prompt and clip-splitting protocol~\cite{yang2026longlive}.

\textbf{Computational cost.} NFE denotes the number of function evaluations in the source sampler~\cite{sabour2024ays}; the SDXL panel retains DMD2's network-forward-pass accounting~\cite{yin2024improved}. Generator GFLOPs and parameter counts describe the cost of each evaluation~\cite{geng2026improvedmeanflow}. For video streaming, throughput measures generated frames per second, and first-frame latency measures the initial response time~\cite{huang2025selfforcing}. Panels (e) and (f) use the single-H100 measurements reported in their respective studies.

\subsection{Single-Evaluation Image Generation}
Table~\ref{tab:quantitative}(a) follows the class-conditional ImageNet $256\times256$ comparison in improved MeanFlow (iMF)~\cite{geng2026improvedmeanflow}. FID is computed on 50K generated images, and generator GFLOPs exclude the tokenizer decoder. All eight configurations use one network evaluation.

iMF supervises instantaneous velocity while retaining the average-velocity update at inference. At similar per-pass cost, iMF-M/2 achieves FID 2.27 at 49.9 GFLOPs and MF-M/2 achieves 5.01 at 54.0 GFLOPs. Scaling iMF from L/2 to XL/2 improves FID from 1.86 to 1.72 at one evaluation, with per-pass cost rising from 116.4 to 174.6 GFLOPs.

\subsection{Text-to-Image Distillation}
Table~\ref{tab:quantitative}(b) uses DMD2's SDXL evaluation on 10K COCO 2014 validation prompts~\cite{yin2024improved}. Global FID compares generated images resized to \(512\times512\) with the corresponding 10K real images.

Among the single-pass models, DMD2 has the lowest global FID (19.01) and SDXL-Turbo the highest CLIP score (0.337). Moving to four DMD2 passes reduces patch FID by 6.12 points, from 26.98 to 20.86, while global FID changes from 19.01 to 19.32. Four-pass DMD2 also matches the guided SDXL teacher (CFG 6) on CLIP alignment at 0.332, with four forward passes instead of 100.

\subsection{Solver and Schedule Interaction}
Table~\ref{tab:quantitative}(c) compares timestep schedules on an ImageNet \(256\times256\) latent diffusion model with CFG 2.0, using 50K samples per setting~\cite{sabour2024ays}.

AYS redistributes timesteps for the chosen sampler, so its effect varies with both the update rule and the budget. At NFE 10, replacing EDM spacing with AYS lowers FID from 66.71 to 23.13 for stochastic DDIM and from 8.48 to 6.11 for SDE-DPM-Solver++(2M). Against time-uniform spacing, AYS improves FID from 8.47 to 6.11 for the latter solver at NFE 10, while at NFE 30 time-uniform spacing scores 7.14 and AYS scores 7.28.

For deterministic DPM-Solver++(2M) at NFE 20, AYS lowers FID from 4.81 to 4.70 and sFID from 5.41 to 5.34 relative to LogSNR, while IS falls from 287.20 to 284.17.

\subsection{Caching across Video Backbones}
Table~\ref{tab:quantitative-video}(d) compares cache policies on Wan 2.1, CogVideoX, and LTX-Video using the full VBench prompt set and a 50-step schedule for each backbone~\cite{haghighi2026sencache}.

MagCache and SenCache tie on LPIPS and SSIM at 25 Wan 2.1 evaluations. At 21 evaluations, SenCache lowers LPIPS from 0.0603 to 0.0540 and raises PSNR from 28.3684 to 29.1400 relative to MagCache. On CogVideoX at 22 evaluations, SenCache gives LPIPS 0.1901 and SSIM 0.7786, compared with 0.5855 and 0.5702 for TeaCache. SenCache estimates sensitivity to changes in both the latent and timestep when selecting refresh points. On LTX-Video, it reaches LPIPS 0.1625 at 27 evaluations versus 0.1795 at 28 for MagCache.

\subsection{Responsiveness and Long-Video Generation}
Chunk-wise Self Forcing and the official Wan-1.3B CausVid implementation both reach 17.0 FPS with 0.69 s first-frame latency~\cite{huang2025selfforcing}. Self Forcing conditions on its own generated history during training, aligning the training rollout with streaming inference. The largest score difference in Table~\ref{tab:quantitative-video}(e) is semantic alignment, at 81.28 for chunk-wise Self Forcing and 69.80 for CausVid.

Frame-wise Self Forcing can return its first frame without waiting for a whole chunk, cutting startup by 0.24 s (0.69 to 0.45 s). Processing each frame separately reduces throughput from 17.0 to 8.9 FPS. Chunk-wise generation is the variant that sustains 16 FPS playback.

LongLive trains on long rollouts and keeps streaming attention bounded with a local key--value window and a frame sink~\cite{yang2026longlive}. On 30-second VBench-Long clips (Table~\ref{tab:quantitative-video}(f)), it reaches 20.7 FPS and scores 83.52 in total, ahead of Self Forcing at 17.0 FPS and 81.59. FramePack scores 81.95 at 0.92 FPS, and SkyReels-V2 scores 75.29 at 0.49 FPS.

\section{Future Directions}
\label{sec:future}
The results above motivate three directions: pairing visual-native architectures with post-compression distillation, training few-step models for their deployment solver, and combining semantic trajectory targets with factorized distribution matching.

\subsection{Visual-Native Architectures and Post-Compression Distillation}
\label{sec:future-arch}

\par\smallskip
\noindent\textbf{Visual Hybrid Architectures.}
Image and video generation place different demands on sequence mixers than language modeling. Nearby patches often share edges or textures, and neighboring frames often depict the same objects in motion. Local convolutions can preserve this structure without loading every interaction into attention. Long-range attention still matters for scene layout, object identity, and subjects that leave and re-enter the frame. SANA combines linear attention with local convolutions for images, while SANA-Video uses block-linear attention to handle long video sequences~\cite{xie2024sana,chen2026sanavideo}.
The latency advantage depends on latent token count, the attention kernel, and the GPU. FlashAttention-2 and FlashAttention-3 reduce the cost of exact attention~\cite{dao2024flashattention2,shah2024flashattention3}, so a hybrid design can retain it where global interactions matter and use local or linear mixing elsewhere.

\par\smallskip
\noindent\textbf{Post-Compression Trajectory and Distribution Recovery.}
Post-training quantization and structured pruning reduce per-step computation, but small errors can propagate when the compressed student's output becomes its next input. Quantization-aware methods recover accuracy at low precision~\cite{li2023qdiffusion,li2025svdquant,he2024efficientdm}. A recovery loss on isolated noisy states does not directly constrain the trajectory produced by the compressed model.

Consistency distillation can shorten that trajectory by teaching direct transitions~\cite{song2023consistency}. DMD instead matches the distribution of the student's own rollouts to the teacher~\cite{yin2024one,yin2024improved}. Comparing these objectives with single-step knowledge distillation at the same bit width, sparsity, and evaluation budget would isolate how much quality each recovers.

\par\smallskip
\noindent\textbf{Causal Adaptation as Structured Pruning.}
A causal temporal mask prunes the bidirectional model's lookahead connections, so the student must generate each chunk from its history. Self Forcing warm-starts a causal student on teacher-generated ODE pairs before on-policy DMD~\cite{huang2025selfforcing}. SolarWM instead initializes the student with teacher-forced AnyFlow, which learns finite-interval transitions, before DMD~\cite{huang2026solarwm,gu2026anyflow}. Consistency and MeanFlow objectives offer related ways to learn long transitions directly~\cite{lu2025scm,geng2025meanflow}. A direct comparison should hold the downstream DMD stage fixed and count both data synthesis and optimization time to reach the same long-video quality.
\subsection{From Training-Free Samplers to Solver-Aware Training}
\label{sec:future-sampler}
Training-free ODE/SDE solvers and optimized timestep schedules have yielded substantial gains at moderate budgets (10--30 NFE)~\cite{lu2022dpm1,zhao2024unipc,sabour2024ays}. At 1~NFE there are no intermediate timesteps to optimize. At 2--4~NFE, a fixed denoiser supplies only a few evaluations across the noise range. Rescaling the path and improving the quadrature are useful, but neither trains the denoiser to predict across the long intervals that this budget imposes.

\par\smallskip
\noindent\textbf{Path Geometry and Schedule Rescaling.}
Appendix~B of sCM shows how different Gaussian schedules map to a common parameterization~\cite{lu2025scm}. Let $x_t=\alpha_t x_0+\sigma_t z$, where $x_0$ has per-coordinate variance $\sigma_d^2$ and $z\sim\mathcal{N}(0,\sigma_d^2 I)$ is independent of $x_0$. Write the EDM-preconditioned clean prediction as~\cite{karras2022elucidating}
\begin{equation*}
\begin{aligned}
D_\theta(x_t,t) &= c_{\mathrm{skip}}(t)x_t\\
&\quad + c_{\mathrm{out}}(t)\mathcal{F}_\theta\bigl(c_{\mathrm{in}}(t)x_t,c_{\mathrm{noise}}(t)\bigr).
\end{aligned}
\end{equation*}
Unit input and residual-target variance give
\begin{equation*}
\begin{aligned}
\operatorname{Var}[c_{\mathrm{in}}(t)x_t] &= 1,\\
c_{\mathrm{out}}^2(t)
 &= \sigma_d^2\bigl(1-c_{\mathrm{skip}}(t)\alpha_t\bigr)^2\\
 &\quad + \sigma_d^2 c_{\mathrm{skip}}^2(t)\sigma_t^2 .
\end{aligned}
\end{equation*}
Let $R_t=\sqrt{\alpha_t^2+\sigma_t^2}$. Minimizing $c_{\mathrm{out}}^2(t)$ over $c_{\mathrm{skip}}(t)$ and choosing the TrigFlow sign gives
\begin{equation}
\label{eq:scm-precond}
\begin{aligned}
c_{\mathrm{in}}(t)&=\frac{1}{\sigma_d R_t},&
c_{\mathrm{skip}}(t)&=\frac{\alpha_t}{R_t^2},\\
c_{\mathrm{out}}(t)&=-\frac{\sigma_d\sigma_t}{R_t}.
\end{aligned}
\end{equation}
The assembled predictor is unchanged by the joint rescaling
$(\hat{\alpha}_t,\hat{\sigma}_t,\hat{x}_t)=(\alpha_t,\sigma_t,x_t)/R_t$,
with $\hat{\alpha}_t^2+\hat{\sigma}_t^2=1$ and unchanged noise conditioning.
Write $\mathcal{F}_t=\mathcal{F}_\theta(\hat{x}_t/\sigma_d,c_{\mathrm{noise}}(t))$. The clean prediction and its weighted squared error then take the form
\begin{equation}
\label{eq:scm-obj}
\begin{aligned}
D_\theta(x_t,t)&=\hat D_\theta(\hat{x}_t,t)
 = \hat{\alpha}_t\hat{x}_t-\hat{\sigma}_t\sigma_d\mathcal{F}_t,\\
\mathcal{L}_{\mathrm{Diff}}
&=\mathbb E_{x_0,z,t}\Bigl[w(t)\hat{\sigma}_t^2\\[-0.2em]
&\qquad {}\cdot
 \bigl\|\sigma_d\mathcal{F}_t-(\hat{\alpha}_t z-\hat{\sigma}_t x_0)\bigr\|_2^2\Bigr].
\end{aligned}
\end{equation}
Where $\alpha_t,\sigma_t>0$, the ratio $x_t/\sigma_t$ and the half-log-SNR $\lambda_t=\log(\alpha_t/\sigma_t)$ are also unchanged. For endpoints in this range, the exact DPM-Solver++ integral becomes~\cite{lu2022dpm1}
\begin{equation}
\label{eq:schedule-invariance}
\begin{aligned}
\frac{x_r}{\sigma_r} &= \frac{x_t}{\sigma_t}
 + \int_{\lambda_t}^{\lambda_r} e^\lambda D_\theta(x_\lambda,\lambda)\,\mathrm{d}\lambda,\\
\frac{\hat{x}_r}{\hat{\sigma}_r} &= \frac{\hat{x}_t}{\hat{\sigma}_t}
 + \int_{\lambda_t}^{\lambda_r} e^\lambda
   \hat D_\theta(\hat{x}_\lambda,\lambda)\,\mathrm{d}\lambda .
\end{aligned}
\end{equation}
The two lines describe the same exact transport, even though the path may look straight in one coordinate system and curved in another. Straightening a path by rescaling does not itself remove few-step discretization error. Schedule changes can still affect the training weights in Eq.~\eqref{eq:scm-obj} and the placement of a finite inference grid~\cite{lipman2022flow,lu2022dpm,lu2022dpm1}.

\par\smallskip
\noindent\textbf{Local Truncation Error at Low NFE.}
At a fixed NFE budget, long intervals can amplify discretization error. On $\alpha_t > 0$, rewriting the teacher PF-ODE in noise-to-signal coordinates $\rho_t = \sigma_t/\alpha_t$ and $\bar{x}_t = x_t/\alpha_t$ simplifies the dynamics to $\mathrm{d}\bar{x}/\mathrm{d}\rho = \epsilon_\Phi(x_\rho, \rho)$~\cite{wang2025rectifieddiffusion}. Subtracting a single deterministic DDIM step from the exact solution over $[r, t]$ yields
\begin{equation}
\label{eq:ddim-error}
\begin{aligned}
\bar{x}_r-\bar{x}_r^{\mathrm{DDIM}}
&= \int_{\rho_t}^{\rho_r}
 \bigl[\epsilon_\Phi(x_\rho,\rho)-\epsilon_\Phi(x_t,\rho_t)\bigr]
 \,\mathrm{d}\rho .
\end{aligned}
\end{equation}
If $\epsilon_\Phi(x_\rho,\rho)$ is $L$-Lipschitz in $\rho$ along the trajectory, the one-step error is at most $\tfrac{L}{2}|\rho_r-\rho_t|^2$. It vanishes on every subinterval when the denoiser output is constant along the path. This condition is about the variation of the integrand, not whether the trajectory is straight in the original coordinates~\cite{wang2025rectifieddiffusion}. A $p$-th-order solver has local error $\mathcal{O}(|\rho_r-\rho_t|^{p+1})$ under sufficient smoothness. At 1--4~NFE the intervals are long, so the small-step error order alone says little about how well a solver recovers the finite transition.

\par\smallskip
\noindent\textbf{Solver-Aware Training Targets.}
On ImageNet $512\times512$, the teacher of sCM runs a 32-step DPM-Solver-2S (Heun) sampler with classifier-free guidance ($63 \times 2$ NFE) to achieve FID~$1.73$, whereas the distilled sCD-XXL absorbs guidance conditioning and predicts endpoints directly, reaching FID~$1.88$ in just 2~NFE (one endpoint prediction, intermediate re-noising to $t=1.1$, and a second endpoint prediction)~\cite{lu2025scm}. A solver-aware objective can train the student to predict the average direction over a finite interval:
\begin{equation*}
\mathcal{T}_{t \to r}(x_t) \approx \frac{1}{\rho_r - \rho_t}\int_{\rho_t}^{\rho_r}\epsilon_\Phi(x_\rho, \rho)\,\mathrm{d}\rho
\end{equation*}
MeanFlow and AnyFlow learn interval-average velocities or flow maps~\cite{geng2025meanflow,gu2026anyflow}. Progressive Distillation and Rectified Diffusion instead supervise teacher endpoints~\cite{salimans2022progressive,wang2025rectifieddiffusion}. At one to four evaluations, training can prioritize the intervals on the deployment grid and expose later updates to states the student actually produces.

\subsection{From Semantic Trajectory Alignment to Factorized Distribution Distillation}
\label{sec:future-distill}
Few-step distillation has to preserve both the structure of individual samples and diversity across seeds while taking longer steps. Semantic targets address the first problem. Distribution matching addresses the second by comparing student and target samples beyond individual teacher--student pairs.

\par\smallskip
\noindent\textbf{High-Noise Pixel Regression and Semantic Alignment.}
Under the rescaled diffusion coordinate $u = x_t/\alpha_t = x_0 + h\epsilon$ (for $\epsilon\sim\mathcal{N}(0,I)$ and bandwidth $h=\sigma_t/\alpha_t$), directly regressing the student's clean prediction $\hat{x}_0(x_t,t,c)$ against ground-truth pixels $x_0$ yields the Bayes-optimal target
\begin{equation}
\label{eq:mean-attraction}
\mu_{p,h}(u) = \mathbb{E}_{x_0 \sim p}[x_0 \mid u, c]
\xrightarrow{\;h \to \infty\;}
\mathbb{E}[x_0 \mid c].
\end{equation}
At high noise, different layouts or poses can yield similar noisy observations. A pixel-space posterior mean averages those alternatives and weakens the link between the seed and the composition. Multi-step teachers restore a seed-specific target through pairs \((z,x_0^\Phi(z))\)~\cite{salimans2022progressive,wang2025rectifieddiffusion}. Even with these pairs, pixel MSE can penalize a small spatial shift as strongly as a change in structure.
Aligning student and teacher predictions through a frozen semantic encoder $\psi(\cdot)$ (such as DINOv2~\cite{oquab2023dinov2}) changes which differences drive the regression gradient:
\begin{equation}
\label{eq:semantic-precond}
\begin{aligned}
\nabla_{\hat{x}_0} \bigl\|\psi(\hat{x}_0)-\psi(x_0^\Phi)\bigr\|_2^2
&= 2J_\psi(\hat{x}_0)^\top \\[-0.25em]
&\quad {}\cdot\bigl(\psi(\hat{x}_0)-\psi(x_0^\Phi)\bigr).
\end{aligned}
\end{equation}
The encoder Jacobian maps feature residuals back to image space. A feature encoder sensitive to object identity and layout can emphasize these differences over small texture shifts, leaving later steps to refine texture and fine motion. Recent methods illustrate different ways to assign that supervision: Align Your Trajectory Tangent stabilizes consistency training via manifold-aligned feature distances~\cite{kim2026ayt}; SwiftVideo combines consistency distillation with frozen DINOv2 features and spatiotemporal discriminator heads~\cite{sun2026swiftvideo}; and DP-DMD supervises the first step with teacher trajectories before applying distribution matching to subsequent refinement steps, thereby preserving seed diversity~\cite{wu2026dpdmd}.

\par\smallskip
\noindent\textbf{Attraction and Repulsion in Score Distillation.}
Semantic alignment stabilizes paired trajectory learning, yet Eq.~\eqref{eq:mean-attraction} reveals a deeper limitation of any pure regression objective: supervising against the teacher posterior mean $\mu_{p,h}(u)$ produces only an \emph{attractive} force toward high-density regions of the smoothed target $p_h = p * \mathcal{N}(0,h^2 I)$, with no countervailing force preventing student samples from crowding onto the same mean. To match distributions directly without paired endpoints, one must subtract the student's own smoothed posterior mean $\mu_{q,h}(u) = \mathbb{E}_{x_0 \sim q}[x_0 \mid u, c]$ at the same bandwidth $h$. Applying Tweedie's identity, $\nabla_u \log p_h(u) = h^{-2}(\mu_{p,h}(u) - u)$ and $\nabla_u \log q_h(u) = h^{-2}(\mu_{q,h}(u) - u)$, cancels the state $u$. With the same Gaussian kernel for $p$ and $q$, the population drifting field~\cite{deng2026drifting,lai2026unifieddrifting,zhu2026mbdmd} is proportional to the reverse-KL score-distillation direction~\cite{luo2023diff,yin2024one,yin2024improved}:
\begin{equation}
\label{eq:drifting-equiv}
\begin{aligned}
V_h(u) &= \mu_{p,h}(u) - \mu_{q,h}(u)
= h^2 \nabla_u \log \frac{p_h(u)}{q_h(u)} \\
&= \frac{\sigma_t^2}{\alpha_t}\bigl(s_{p_t}(y_t) - s_{q_t}(y_t)\bigr),
\end{aligned}
\end{equation}
where $y_t = \alpha_t u$, $h = \sigma_t/\alpha_t$, and the last equality follows from the chain rule $\nabla_u \log p_h(u) = \alpha_t s_{p_t}(y_t)$. The teacher term \(+s_{p_t}\) attracts samples toward high-density regions of the target. The student term \(-s_{q_t}\) pushes back where the student distribution crowds. The result depends on how well the auxiliary score tracks the student and on which marginals are compared.

\par\smallskip
\noindent\textbf{Fake-Score Bias and Mode Coverage.}
An inaccurate auxiliary score changes the repulsive term in Eq.~\eqref{eq:drifting-equiv}. A two-mode example shows how this can alter local stability. Consider a symmetric bimodal target $p = \frac{1}{2}\delta_{-a} + \frac{1}{2}\delta_{+a}$ and a centered Gaussian student $q = \mathcal{N}(0, B)$, both smoothed at bandwidth $h$. The smoothed densities yield exact scores $s_p(u) = -\frac{u}{h^2} + \frac{a}{h^2}\tanh(au/h^2)$ and $s_q(u) = -\frac{u}{B+h^2}$. If an under-fitted or lagging fake model locally approximates the student score as $\gamma s_q(u)$ ($\gamma > 0$), differentiating the resulting update field $g_\gamma(u) = s_p(u) - \gamma s_q(u)$ at the midpoint $u=0$ yields
\begin{equation}
\label{eq:fake-score-slope}
g_\gamma'(0)
= \frac{a^2}{h^4} - \frac{1}{h^2} + \frac{\gamma}{B + h^2}.
\end{equation}
A positive slope pushes samples near the midpoint away from it, whereas a negative slope draws them back. Setting $a^2 = B = h^2/2$ with fixed teacher and bandwidth gives $g_\gamma'(0) = \frac{1}{h^2}\bigl(\frac{2}{3}\gamma - \frac{1}{2}\bigr)$; as soon as $\gamma$ drops below $3/4$, the slope flips sign. In this example, bandwidth $h$ sets the scale at which the modes interact, while $\gamma$ changes the local strength of the anti-crowding term. DMD2 addresses a related practical issue by updating the fake denoiser more frequently so it tracks the student's evolving distribution~\cite{yin2024improved}. Fidelity metrics can also move in different directions: in sCM's ImageNet $512\times512$ ablation, adding adaptive VSD to sCD-M improves two-step DINOv2 Fr\'echet distance from $55.70$ to $53.53$, while FID shifts slightly from $2.26$ to $2.29$~\cite{lu2025scm}. Consequently, future distillation evaluations should vary $h$ and fake-model capacity/update ratio independently while jointly tracking per-prompt seed diversity, recall, and FID.

\par\smallskip
\noindent\textbf{Factorized Distribution Matching for Video.}
Single-frame marginals capture appearance but cannot distinguish a clip from a permutation of its frames. A joint video critic can capture temporal order, but it need not provide the strongest supervision for frame detail. The two signals can be combined with a spatial mask:
\begin{equation}
\label{eq:factorized-dmd}
\begin{aligned}
\Delta_{\mathrm{video}}
&= w_{\mathrm{vid}}\bigl(s_{p}^{\mathrm{vid}} - s_{q}^{\mathrm{vid}}\bigr) \\
&\quad + w_{\mathrm{img}}\, M_{\mathrm{spatial}} \odot
\bigl(s_{p}^{\mathrm{img}} - s_{q}^{\mathrm{img}}\bigr),
\end{aligned}
\end{equation}
where $(s_p^{\mathrm{vid}}, s_q^{\mathrm{vid}})$ constrain joint spatiotemporal realism, $(s_p^{\mathrm{img}},s_q^{\mathrm{img}})$ compare single-frame appearance through a separate image branch, and $M_{\mathrm{spatial}}$ controls where this branch acts. When the two latent spaces differ, let \(B\) map a video latent to an image-frame latent. \(B\) can decode the video latent, select a frame, and re-encode it with an image VAE, or it can be a learned latent projection. The image score difference in Eq.~\eqref{eq:factorized-dmd} denotes the image-space difference pulled back by \(J_B^\top\) before masking. DuoMatching adopts this joint-plus-marginal decomposition ($M_{\mathrm{spatial}} \equiv 1$) to strengthen few-step video distillation~\cite{zhan2026duomatching}, whereas LynnReal-Hand pairs a hand- and camera-conditioned video branch acting over the full clip with an image branch whose gradients are masked to non-hand regions~\cite{rui2026lynnrealhand}. On the full-hand subset of HandBench under a three-step sampling setting, adding the image branch reduces FVD from $289.05$ to $208.75$; further masking its gradients spatially away from the hand lowers FVD to $198.99$ while raising MANO Dice from $92.00$ to $92.50$~\cite{rui2026lynnrealhand}. For interactive world models, the next test is whether separate appearance and action-conditioned critics can improve frame detail without narrowing the range of motions at the same NFE.

\section{Conclusion}
\label{sec:conclusion}
Distillation and consistency training make one- to four-pass image synthesis possible by learning larger transitions. Improved MeanFlow reaches FID 1.72 on ImageNet \(256\times256\) in one evaluation~\cite{geng2025meanflow,geng2026improvedmeanflow}. The SDXL and ImageNet results in Table~\ref{tab:quantitative} show two effects of the evaluation budget. Extra DMD2 passes improve patch FID while global FID changes little, and the FID ranking of time-uniform and AYS schedules can reverse when more steps are available.

In video, selective refresh and feature prediction reduce repeated denoiser work~\cite{liu2025taylorseer,haghighi2026sencache}. Frame-wise Self Forcing returns its first frame sooner, while chunk-wise decoding sustains 17.0 FPS. LongLive reaches 20.7 FPS on 30-second clips using a bounded attention window and a frame sink~\cite{huang2025selfforcing,yang2026longlive}.

The remaining questions concern how these gains combine. Visual backbones can retain local structure while reserving global attention for long-range interactions, and post-compression distillation can target the errors that accumulate during rollout. At one to four evaluations, training for the deployed solver intervals addresses errors that schedule changes alone do not remove. For video, semantic targets can preserve layout during large transitions, while separate frame and clip critics let appearance and motion be evaluated independently.

\clearpage
\ifCLASSOPTIONcompsoc
{
	\small
\setlength{\baselineskip}{9.75pt}
\bibliographystyle{ieeetr}
\bibliography{paper}
}
\fi

\ifCLASSOPTIONcaptionsoff
  \newpage
\fi

\end{document}